\pdfoutput=1

\documentclass[11pt]{article}

\usepackage[preprint]{acl}

\usepackage{times}
\usepackage{latexsym}
\usepackage{booktabs}
\usepackage{multirow}
\usepackage{amsmath}
\usepackage{amssymb}
\usepackage{siunitx}
\usepackage{pifont}
\newcommand{\cmark}{\ding{51}} 
\newcommand{\xmark}{\ding{55}} 

\usepackage[T1]{fontenc}

\usepackage[utf8]{inputenc}

\usepackage{microtype}

\usepackage{inconsolata}

\usepackage{graphicx}
\usepackage{subcaption}

\title{DoMIX: An Efficient Framework for Exploiting Domain Knowledge in Fine-Tuning}

\author{
Dohoon Kim$^{1}$ \qquad
    Donghun Kang$^{1}$ \qquad
    Taesup Moon$^{1,2}$\thanks{Corresponding author}
    \\
    $^{1}$ Department of Electrical and Computer Engineering, Seoul National University \\
    $^{2}$ ASRI/INMC/IPAI/AIIS, Seoul National University \\
    {\tt\small \{dohoon.kim, 0k9d0h1, tsmoon\}@snu.ac.kr }
}

\begin{document}
\maketitle
\begin{abstract}
Domain-Adaptive Pre-training (DAP) has recently gained attention for its effectiveness in fine-tuning pre-trained models. Building on this, continual DAP has been explored to develop pre-trained models capable of incrementally incorporating different domain datasets. However, existing continual DAP methods face several limitations: (1) high computational cost and GPU memory usage during training; (2) sensitivity to incremental data order; and (3) providing a single, generalized model for all end tasks, which contradicts the essence of DAP. In this paper, we propose DoMIX, a novel approach that addresses these challenges by leveraging LoRA modules, a representative parameter-efficient fine-tuning (PEFT) method. Our approach enables efficient and parallel domain-adaptive pre-training that is robust to domain order and effectively utilizes accumulated knowledge to provide tailored pre-trained models for specific tasks.
We also demonstrate that our method can be extended beyond the DAP setting to standard LLM fine-tuning scenarios. Code is available at \href{https://github.com/dohoonkim-ai/DoMIX}{\tt{https://github.com/dohoonkim-ai/DoMIX}}.

\end{abstract}
\section{Introduction}
Large Language Models (LLMs) \citep{devlin-etal-2019-bert, liu2019roberta, gpt3, touvron2023llama, achiam2023gpt, team2024gemma, llama3modelcard} have demonstrated exceptional performance across various tasks, including sentiment classification \citep{devlin-etal-2019-bert, liu2019roberta}, commonsense reasoning \citep{sap2020commonsense}, arithmetic reasoning \cite{gpt3, wei2022chain}, and natural language understanding \citep{wang2018glue}. This success is attributed to a training strategy involving two steps—pre-training and fine-tuning—which has now become standard practice. 

Moreover, instead of directly fine-tuning on the end task, conducting additional pre-training on unlabeled domain-specific datasets related to the end task, followed by fine-tuning on the end-task datasets, has been shown to be far more effective \citep{gururangan2020don, xu-etal-2019-bertpost, DGA}. This training framework is known as Domain-Adaptive Pre-training (DAP).
Furthermore, \citet{CTR, ke2021classic, keBCL, CPT, DGA, ke2023continual} have extended the DAP setting to a continual DAP setting, arguing that in real-world scenarios, there is a need for a framework that incrementally incorporates knowledge from newly obtained domain datasets into a foundation model.

However, existing continual DAP methods \citep{CTR, ke2021classic, keBCL, CPT, DGA, ke2023continual} and traditional continual learning methods \citep{EWC, serra2018hat, buzzega2020dark}, which can be applied to continual DAP, lack efficiency in domain knowledge accumulation, incurring high computational costs due to the need to prevent catastrophic forgetting while integrating new domain knowledge. This inefficiency hinders their application in real-world scenarios, especially when managing large datasets. Moreover, these methods are limited to settings where datasets arrive incrementally, whereas real-world data are often obtained simultaneously through multiple channels. Due to the nature of sequential methods, where training on new data depends on previously processed data, they are sensitive to the order in which data are processed, as we will show in Section~\ref{sec:DAPResults}. Therefore, we argue that a new training framework is needed to enable efficient, parallel accumulation of domain knowledge.



Furthermore, the continual DAP methods \citep{EWC, serra2018hat, buzzega2020dark, CTR, ke2021classic, keBCL, CPT, DGA, ke2023continual} lack the ability to exploit appropriate domain knowledge for a specific task. This is because they accumulate all domain knowledge, especially general knowledge, into a single model and fine-tune this model to any target task. The essence of DAP is to provide appropriate model for each target task but continual DAP does not satisfy the essence by providing a same model for any target task. Knowledge contained within unlabeled domain corpora is valuable not only for its general natural language understanding but also for its specialized, domain-specific knowledge. However, existing continual DAP methods compromise the specialization of each domain by merging them into a single model. 



As a result, we propose a new training framework based on LoRA \citep{hu2021lora}, a representative parameter-efficient fine-tuning (PEFT) method. For knowledge accumulation, we preserve the specialization of each domain by separately storing domain knowledge in distinct LoRA modules. For knowledge exploitation, we concatenate the domain-specific LoRA modules at each layer and introduce a square bridge module—diagonally initialized for effective integration—while partially freezing the LoRA modules. This design efficiently and effectively leverages the accumulated domain knowledge for the end task. Our contributions are as follows:


\begin{itemize}
\item We propose a novel knowledge accumulation and exploitation framework based on LoRA, where domain knowledge is saved separately and flexibly exploited for target tasks.
\item In the continual DAP scenario, we demonstrate superior performance compared to state-of-the-art continual DAP methods, while reducing pre-training time by 58\%, GPU memory usage during pre-training by 87\%, and GPU memory usage during end-task fine-tuning by 37\%.
\item We also extend our method to a standard LLM fine-tuning scenario, achieving superior performance while reducing training time by 36\% and memory usage by 18\% compared to the state-of-the-art PEFT method.
\end{itemize}


\section{Related Work}

\textbf{Continual Domain-Adaptive Pre-training.} 
Performing pre-training on domain-specific datasets related to the end task—a process known as domain-adaptive pre-training (DAP)—has proven effective \citep{alsentzer2019publicly, gururangan2020don, lee2020biobert, xu-etal-2019-bertpost, sun2019fine}. Building on this idea, the continual DAP setting has emerged, applying the concept of continual learning (CL) \citep{EWC, serra2018hat, buzzega2020dark, liang2024inflora} to DAP, with the goal of constructing a single model that incrementally incorporates domain knowledge \citep{DGA, ke2023continual}. However, previous work often underestimates the computational and memory overheads required to prevent catastrophic forgetting and preserve general knowledge, thereby limiting real-world applicability. In contrast, we improve the efficiency of both pre-training and fine-tuning, reducing training time and GPU memory usage without compromising performance. Our method builds on insights from \citet{liang2024inflora}, which demonstrated that freezing specific weights constrains model updates to the corresponding subspace, as detailed in Section~\ref{section:method:finetuning}.

%


\noindent\textbf{Parameter-Efficient Fine-Tuning.}
Parameter-Efficient Fine-Tuning (PEFT) \citep{AdapterH, pfeiffer2020madadpaterP, he2021towardsParallelAdapter, Prompttuning, Prefixtuning, hu2021lora} has emerged as a promising approach to fine-tune LLMs with significantly reduced memory and computational costs. In this paper, we focus on LoRA \citep{hu2021lora}, which has become a standard PEFT method. Several LoRA variants have been proposed to improve its effectiveness \citep{liu2024dora, si2024unleashing, wu2024moslora, kopiczko2024vera}. Our method shares similarities with \citet{wu2024moslora} in introducing additional modules within the LoRA architecture, but differs in initialization strategy and overall objective, as discussed in Section~\ref{section:ablation:effect of Initialization}.




\begin{figure}[t]
  \includegraphics[width=\columnwidth]{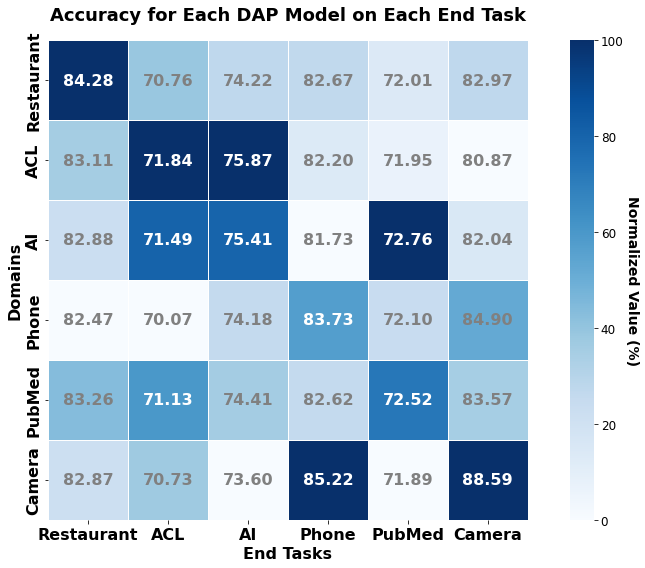}
  \caption{Cross-domain transferability of domain-adaptive pre-training. Each cell $(i, j)$ shows the end-task performance when using domain-adaptive pre-training on domain $i$ and fine-tuning on task $j$.}
  \label{fig:MotivationHeatmap}
\end{figure}




\section{Motivation}

In this section, we present our observations on Domain-Adaptive Pre-training (DAP) that motivate our proposed method, DoMIX.

Following the dataset configuration of \citet{ke2023continual}, we consider 6 domain datasets — \textit{Yelp Restaurant} \citep{xu-etal-2019-bertpost}, \textit{Amazon Phone} \citep{ni2019justifying}, \textit{Amazon Camera} \citep{ni2019justifying}, \textit{ACL Papers} \citep{lo2019s2orc}, \textit{AI Papers} \citep{lo2019s2orc}, and \textit{PubMed Papers}\footnote{\url{https://pubmed.ncbi.nlm.nih.gov/}} — and their corresponding 6 end tasks. Further details are provided in Appendix~\ref{section:datasets:dap datasets}.
Starting from the RoBERTa-Base model \citep{liu2019roberta}, we adapt it to each of the 6 domain datasets using LoRA \citep{hu2021lora}, resulting in 6 separate domain-adapted models. We then fully fine-tune each of these models on the corresponding end tasks, mirroring the evaluation process of \citet{ke2023continual}. Figure~\ref{fig:MotivationHeatmap} illustrates the results;
the rows represent domain-adaptive pre-trained models trained on each domain corpus, while the columns correspond to the respective end-task datasets. 
Each entry \((i, j)\) denotes the performance of the model pre-trained on domain \(i\) when evaluated on task \(j\). 
For reference, each domain corresponds to its respective task in the same order (e.g., Restaurant domain with Restaurant task, and so on).
Further details of this experimental setup are provided in Appendix~\ref{section: Appendix: DetailofMotivation}.

Previous DAP studies \citep{gururangan2020don, xu-etal-2019-bertpost, DGA} commonly assume that domain knowledge related to a given end task is most beneficial for that specific task. Under this assumption, each diagonal entry in our results should surpass all other entries in the corresponding column. However, our experiments indicate that this is not always the case—domain knowledge from different domains can also be helpful. For instance, in the AI, Phone, and PubMed tasks, we observe such behavior.

As a result, we argue that certain domain knowledge can indeed be beneficial for specific end tasks, but it is often unclear which domain will be most helpful for a given task. Therefore, we propose a novel framework that allows models to flexibly exploit pre-trained domain knowledge during fine-tuning, regardless of the target task.

\begin{figure*}[t]
  \includegraphics[width=\textwidth]{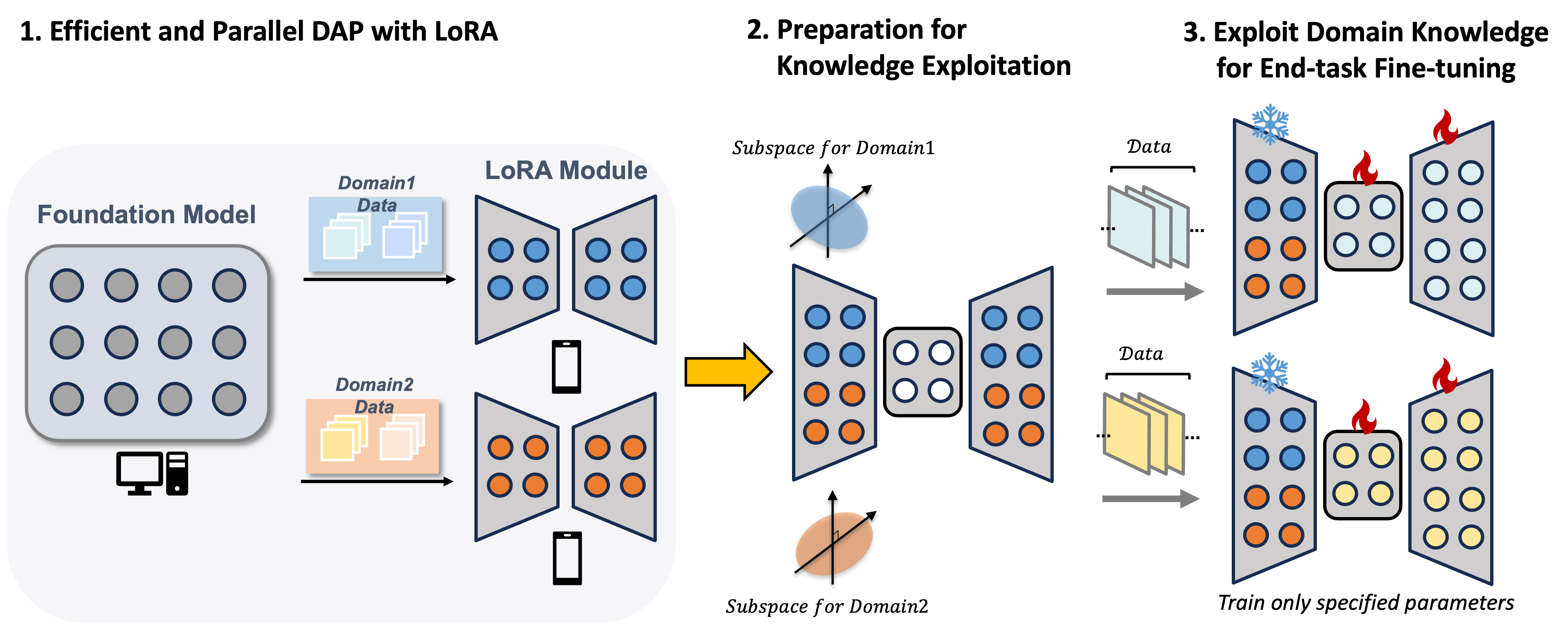}
  \caption{Overall Framework of DoMIX.}
  \label{fig:overallmethodframwork}
\end{figure*}

\section{Method}
In this section, we present the preliminaries and describe our method, DoMIX, in detail. The overall framework is illustrated in Figure~\ref{fig:overallmethodframwork}.

\subsection{Preliminaries}
In this section, we briefly explain LoRA \citep{hu2021lora}, a PEFT method that adapts pre-trained models by introducing low-rank learnable matrices to approximate full weight updates. LoRA keeps the original model weights frozen and updates only the residual weights, represented as:
\[
W' = W + \Delta W, \quad \Delta W = BA,
\]
in which \( W \in \mathbb{R}^{m \times n} \) represents the frozen original model weights, \( \Delta W \) is the learnable residual weight, and \( A \in \mathbb{R}^{r \times n} \) and \( B \in \mathbb{R}^{m \times r} \) are low-rank matrices with rank \( r \ll \min(m, n) \). By decomposing \( \Delta W \) into the product of \( A \) and \( B \), LoRA significantly reduces the number of trainable parameters while maintaining competitive performance.

\subsection{Three Steps of DoMIX}
\noindent{\textbf{(1) Efficient and Parallel DAP with LoRA.}}\  For each domain dataset, we train a separate LoRA module on the corresponding corpus while keeping the foundation model weights frozen. Specifically, we optimize the standard masked language modeling (MLM) objective:

\begin{equation}
    \mathcal{L}_{\text{MLM}}(\theta) = -\sum_{x \in \mathcal{D}} \sum_{t \in \mathcal{M}(x)} \log p_\theta(x_t \mid x_{\backslash \mathcal{M}(x)}),\nonumber
\end{equation}
where \(\mathcal{D}\) is the training corpus, \(\mathcal{M}(x)\) denotes the set of masked positions in the tokenized sequence \(x\), and \(p_\theta(x_t \mid x_{\backslash \mathcal{M}(x)})\) is the predicted probability of the original token \(x_t\) given the unmasked context.

Notably, we do not need to record any domain identifiers at this stage, as our method is designed to exploit accumulated domain knowledge without relying on explicit domain identification.\vspace{.1in}

\noindent{\textbf{(2) Preparation for Knowledge Exploitation.}}\ 
To exploit the domain knowledge accumulated in the LoRA modules, we propose the following method, as illustrated in Figure~\ref{fig:overallmethodframwork}. Inspired by \citet{wu2024moslora}, the multiplication of LoRA matrices \(A\) and \(B\) can be interpreted as a sum of \(r\) submatrices.
\begin{align}
    \Delta W &= BA \nonumber \\
    &= 
    \begin{bmatrix} 
        | & | & & | \\ 
        b_1 & b_2 & \dots & b_r \\ 
        | & | & & | 
    \end{bmatrix}  
    \begin{bmatrix} 
        -a_1^T- \\ 
        -a_2^T- \\ 
        \vdots \\ 
        -a_r^T- 
    \end{bmatrix} \nonumber \\
    &= b_1a_1^T + b_2a_2^T + \dots + b_ra_r^T\nonumber
\end{align}

\noindent where \( b_i \in \mathbb{R}^m \) is the \( i \)-th column vector of \( B \), and \( a_i^T \in \mathbb{R}^{1 \times n} \) is the \( i \)-th row of \( A \).

In this interpretation, we assume that domain knowledge is encoded in these \( r \) subspaces. Based on this observation, the LoRA modules are concatenated to form unified knowledge subspaces. To flexibly exploit these subspaces, we introduce a diagonally initialized matrix \( P \), referred to as the \textit{bridge module}, between all \( A \) modules (closer to the input) and \( B \) modules (closer to the output). This bridge module serves to control the extent to which each knowledge subspace is emphasized or suppressed. For example, when exploiting two domains of knowledge (represented by $B_1A_1$ and $B_2A_2$, where $B_1, B_2 \in \mathbb{R}^{m \times r}$ and $A_1, A_2 \in \mathbb{R}^{r \times n}$), the corresponding LoRA modules are concatenated in a column-wise manner for $B$ and a row-wise manner for $A$, resulting in $B_{\text{cat}} = [B_1, B_2]$ and $A_{\text{cat}} = \begin{bmatrix} A_1 \\ A_2 \end{bmatrix}$. A diagonally initialized bridge matrix $P$ is then inserted between them. As a result, the weight update equation takes the following form:

\begin{align}
    &\Delta W = B_{\text{cat}} P A_{\text{cat}} \nonumber \\
    &= [B_1\;\; B_2] \; P \; \begin{bmatrix} A_1 \\ A_2 \end{bmatrix} \nonumber \\
    &=
    \scriptsize
    \begin{bmatrix} 
        | & & | \\ 
        b_1 & \dots & b_{2r} \\ 
        | & & | 
    \end{bmatrix} 
    \begin{bmatrix} 
        p_{11} & 0 & \dots & 0 \\ 
        0 & p_{22} & \dots & 0 \\ 
        \vdots & \vdots & \ddots & \vdots \\ 
        0 & 0 & \dots & p_{2r2r} 
    \end{bmatrix} 
    \begin{bmatrix} 
        -a_1^T- \\ 
        -a_2^T- \\ 
        \vdots \\ 
        -a_{2r}^T- 
    \end{bmatrix} \normalsize \nonumber \\
    &= p_{11} b_1a_1^T + p_{22}b_2a_2^T + \dots + p_{2r2r}b_{2r}a_{2r}^T \nonumber
\end{align}

\noindent
where \( b_i \in \mathbb{R}^m \) is the \( i \)-th column vector of \( B_{\text{cat}} \), and \( a_i^T \in \mathbb{R}^{1 \times n} \) is the \( i \)-th row of \( A_{\text{cat}} \). The scalar \( p_{ii} \in \mathbb{R} \) denotes the \( i \)-th diagonal entry of the bridge matrix \( P \in \mathbb{R}^{2r \times 2r} \).

As shown in the equation, the diagonal entries \(p_{ii}\) of \(P\) correspond to the extent of domain-specific knowledge employed. Specifically, entries from \(p_{11}\) to \(p_{rr}\) correspond to the first domain knowledge, while entries from \(p_{r+1,r+1}\) to \(p_{2r2r}\) correspond to the second domain knowledge. \vspace{.1in}

\noindent{\textbf{(3) Exploit Domain Knowledge for End-task Fine-tuning.}} \ \ 
\label{section:method:finetuning}
During fine-tuning on the end task, the \(A\) modules remain frozen, while the bridge matrix $P$ and \(B\) modules are trainable.
InfLoRA \citep{liang2024inflora} demonstrated that freezing the \(A\) modules and training only the \(B\) modules updates the foundation model’s weights within the subspace defined by \(\text{span}(A)\).
Motivated by this finding, we adopt a similar approach by freezing the \(A\) modules. Our objective is to fine-tune the model within the \(A\) subspace, which encodes the relevant domain knowledge. We anticipate that updating parameters exclusively in these subspaces will facilitate more effective exploitation of the accumulated domain knowledge.

We initialize the bridge matrix 
$P$ as a diagonal matrix with uniform diagonal entries that sum to one. By ensuring the diagonal entries are evenly set, we avoid introducing biases regarding which domain knowledge should be favored during fine-tuning. 


As a result, the combination of the bridge matrix and the freezing strategy (freezing the \(A\) modules) enables flexible utilization of accumulated domain knowledge. Specifically, it determines the extent of domain-specific knowledge employed while effectively leveraging domain subspaces.

\begin{table*}[h!]
\caption{Comparison of Accuracy and F1 Scores (micro-F1 for \textit{PubMed}, following \citet{gururangan2020don, dery2021shouldf1, beltagy2019scibertf1}, and macro-F1 for other domains) across different pre-training and fine-tuning methods. \cmark: Indicates that the method requires a domain ID during end-task fine-tuning. Bold values represent the best scores in each column.}
\centering
\resizebox{\textwidth}{!}{
\begin{tabular}{llc|cc cc cc cc c cc|cc}
\toprule
\textbf{Pretrain} & \textbf{Finetune} &\textbf{Domain ID} & \multicolumn{2}{c}{\textbf{Restaurant}} & \multicolumn{2}{c}{\textbf{ACL}} & \multicolumn{2}{c}{\textbf{AI}} & \multicolumn{2}{c}{\textbf{Phone}} & \multicolumn{1}{c}{\textbf{PubMed}} & \multicolumn{2}{c|}{\textbf{Camera}} & \multicolumn{2}{c}{\textbf{Average}} \\
\midrule
& & &Acc & F1 & Acc & F1 & Acc & F1 & Acc & F1 & F1 & Acc & F1 & Acc & F1 \\
\midrule
\multirow{2}{*}{RoBERTa} & Full  & \multirow{2}{*}-     & 86.67 & 79.25 & 72.52 & 68.71 & 76.13 & 70.08 & 86.26 & 83.84  & 72.20 & 89.73 & 83.84 & 80.59 & 76.32 \\
                         & LoRA  &      & 86.17 &	78.60 & 69.52 &	63.33 & 74.15 &	66.51 & 86.94 &	85.15 & 72.40 & 90.70 & 86.10 & 79.98 &	75.35 \\
\midrule
\multirow{2}{*}{Seperate LoRA}    & Full   & \multirow{2}{*}\cmark   & 87.75 & 81.07 & 73.82 & 69.95 & 78.29 & 72.72 & 85.15 & 82.35 & 72.52 & 90.93 & 86.37 & 81.41 & 77.50 \\
                         & LoRA &     & 86.75 &	79.24 & 72.33 &	66.98 & 77.20 & 70.92 & 87.65 &	85.79 & 72.50 & 92.57 &	89.16 & 81.50 &	77.43 \\
\midrule
\multirow{2}{*}{Joint LoRA} & Full & \multirow{2}{*}\xmark   & 87.31 & 80.51 & 73.82 & 69.91 & 78.48 & 72.93 & 84.77 & 81.69 & 72.48 & 89.95 & 84.32 & 81.14 & 76.97 \\
                            & LoRA &    & 86.63 &	79.43 & 71.43 &	66.25 & 76.66 &	69.88 & 85.37 &	82.48 & 72.15 & 91.05 &	86.62 & 80.55 &	76.13 \\
\midrule
\midrule
\multirow{2}{*}{NCL}        & Full & \multirow{2}{*}\xmark   & 87.56 & 80.87 & \textbf{74.74} & \textbf{70.82} & 78.28 & 72.25 & 85.51 & 82.96 & 72.89 & 89.94 & 84.42 & 81.49 & 77.37 \\
                            & LoRA &    & 86.38 &	78.69 & 72.09 &	66.59 & 76.58 &	69.50 & 86.35 &	84.20 & 72.63 & \textbf{93.10} &	\textbf{89.76} & 81.19 &	76.90 \\
\midrule
\multirow{2}{*}{EWC}        & Full & \multirow{2}{*}\xmark & 87.33 & 80.42 & 73.95 & 69.94 & \textbf{79.28} & 73.91 & 85.55 & 82.92 & 72.96 & 89.82 & 83.94 & 81.48 & 77.35   \\
                            & LoRA &  &87.38 &	80.32 & 72.27 &	66.99 & 76.91 &	70.00 &86.43 &	84.37 &72.27 &91.62 &	87.25 & 81.15 &	76.87 \\
\midrule
\multirow{2}{*}{KD}        & Full & \multirow{2}{*}\xmark& 87.27 & 80.41 & 74.31 & 70.11 & 78.20 & 72.33 & 85.99 & 83.69 & 73.04 & 90.96 & 86.08 & 81.63 & 77.61\\
                            & LoRA  & & 87.18 & 	80.21  &72.55 &	67.63 &76.29&	69.19 &86.68&	84.88 & 72.97 &91.78&	87.86 &81.24 &	77.12\\
\midrule

\multirow{2}{*}{DAS}        & Full & \multirow{2}{*}\xmark   & \textbf{87.57} & \textbf{80.90} & 74.47 & 70.52 & 78.72 & 72.85 & 85.40 & 82.82 & 72.99 & 89.98 & 84.47 & 81.52 & 77.43 \\
                            & LoRA &    & 87.44 &	80.62 & 72.02 &	66.37& 76.33 &	68.75 & 86.48 &	84.38 &  72.27 & 91.55 &	87.28 & 81.01 &	76.61 \\
\midrule

Seperate LoRA                        & DoMIX (Ours)  & \xmark  & 86.67 & 79.30 & 72.97 & 69.10 & 79.11 & \textbf{74.01} & \textbf{87.12} & \textbf{85.23} & \textbf{73.61} & 90.54 & 85.79 & \textbf{81.67} & \textbf{77.84} \\
\bottomrule
\end{tabular}
}

\label{tab:DAPResults}
\end{table*}

\begin{table}[ht]
    \centering
    \caption{Comparison of trainable parameters and GPU memory usage during end-task fine-tuning.}
    \resizebox{0.5\textwidth}{!}{
    \label{tab:numparam}
    \begin{tabular}{lcrrr}
        \toprule
        \textbf{Finetune Method} & \textbf{\# of Domains} & \textbf{\# of Params (vs Full)} &\textbf{GPU Mem.} \\
        \midrule
        Full Finetuning & -- & 124.06M (100\%) &6,253MiB \\
        LoRA ($r = 48$) & -- & 11.28M (9.09\%) & 5,076MiB \\
        \midrule
        \multirow{6}{*}{\shortstack[l]{DoMIX (Ours)\\($r = 8$ per domain)}}
        & 1 & 0.67M (0.54\%)&4,063MiB \\
        & 2 & 1.35M (1.09\%)&4,093MiB \\
        & 3 & 2.03M (1.64\%)&4,123MiB \\
        & 4 & 2.73M (2.20\%)&4,173MiB \\
        & 5 & 3.43M (2.76\%)&4,217MiB \\
        & 6 & 4.15M (3.35\%)&4,235MiB \\
        \bottomrule
    \end{tabular}
    }
\end{table}

\begin{figure*}[t]
  \includegraphics[width=\textwidth]{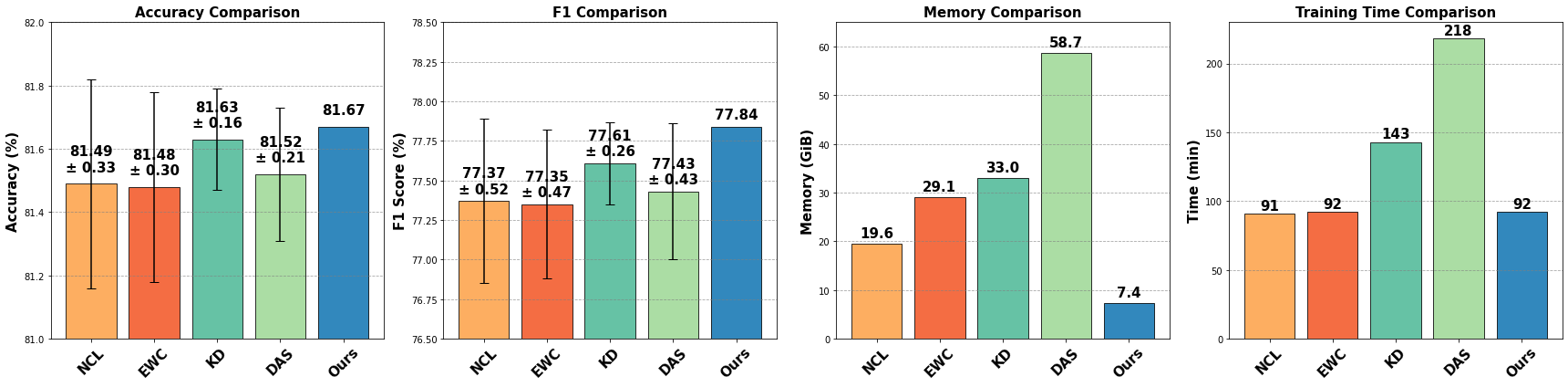}
  \caption{Left: Variance comparison of average accuracy and F1 scores. Right: Memory usage and training time costs during the DAP stage.}
  \label{fig:variancegraph}
\end{figure*}
\section{Experiments}
We demonstrate the efficiency and effectiveness of our method across various experimental settings, including continual DAP scenarios and standard LLM fine-tuning with multiple PEFT methods. In addition, we provide extensive ablation studies to verify the underlying intuitions of our approach.

\subsection{Continual Domain Adaptive Pretraining}
In this setting, we primarily follow the configuration of \citet{ke2023continual}. Specifically, we use RoBERTa-Base \citep{liu2019roberta} as our base model and incrementally train on 6 domains, each associated with a corresponding end task. After this incremental training, we fine-tune the resulting model on each end task. 

Unlike \citet{ke2023continual}, we consider 6 different domain orders to evaluate robustness against variations in data order. Furthermore, when fine-tuning on the end tasks, we explore not only full fine-tuning, but also LoRA-based fine-tuning and our proposed method. End-task fine-tuning is conducted using 10 random seeds for more reliable evaluation. Details of the hyperparameters are provided in Appendix~\ref{section:appendix:DAPDetail}.

\subsubsection{Datasets}
The domain data consist of raw text drawn from several domains: \textit{Yelp Restaurant} \citep{xu-etal-2019-bertpost}, \textit{Amazon Phone} \citep{ni2019justifying}, \textit{Amazon Camera} \citep{ni2019justifying}, \textit{ACL Papers} \citep{lo2019s2orc}, \textit{AI Papers} \citep{lo2019s2orc}, and \textit{PubMed Papers}\footnote{\url{https://pubmed.ncbi.nlm.nih.gov/}}.

There are also six corresponding end-task classification datasets, each containing text and associated labels: \textit{Restaurant}\footnote{\url{https://alt.qcri.org/semeval2014/task4/}}, \textit{Phone} \citep{ding2008holistic, 10.1145/1014052.1014073}, \textit{Camera} \citep{ding2008holistic, 10.1145/1014052.1014073}, \textit{ACL} (ACL-ARC from \citep{jurgens-etal-2018-measuring}), \textit{AI} (SCIERC from \citep{luan2018multi}), and \textit{PubMed} (CHEMPROT from \citep{kringelum2016chemprot}).

\subsubsection{Baselines}

For the domain-adaptive pre-training stage, we compare our method against three Non-CL baselines and four CL baselines. \textbf{RoBERTa} refers to the foundation model without DAP. 
\textbf{Separate LoRA} trains a distinct LoRA module on each domain dataset independently. \textbf{Joint LoRA} trains a single LoRA module across all domain datasets simultaneously.
\textbf{NCL} (Naive Continual Learning) trains a single model as new domain data arrive, without any specialized mechanism for mitigating forgetting. \textbf{EWC} \citep{EWC} is a representative CL method that regularizes changes in parameters deemed important for previously learned domains, thus reducing catastrophic forgetting. \textbf{DAS} \citep{ke2023continual} is a state-of-the-art continual DAP method that employs soft-masking, contrastive loss, and distillation to effectively accumulate domain knowledge without forgetting. \textbf{KD} \citep{hinton2015distilling} applies knowledge distillation whenever new domain data arrive, transferring knowledge from the previous model to the updated model. 

Using the pre-trained models derived from these baselines, we then fine-tune each model on the end tasks with three strategies: full fine-tuning, LoRA-based fine-tuning, and our proposed approach, DoMIX. To elaborate on DoMIX, we first apply \textbf{Separate LoRA} during the domain-adaptive pre-training stage and then utilize the DoMIX framework during end-task fine-tuning.



\subsubsection{Results}
\label{sec:DAPResults}
\noindent\textbf{Performance.} In Table~\ref{tab:DAPResults}, we observe that DoMIX achieves the best average performance. Specifically, ours show the better performance than Seperate LoRA with LoRA finetuning and full-finetuning. This results suggest that our fientuning method exploits several domain knowledges appropriately so that it can improve the performance. 

\noindent\textbf{Efficiency.}
The performance of DoMIX is noteworthy given its efficiency. In the right two plots of Figure~\ref{fig:variancegraph}, we present the peak GPU memory usage and training time during the DAP stage. Our method demonstrates significantly lower memory usage and training time, reducing memory usage by 87\% and training time by 58\% compared to DAS, which is known as a state-of-the-art (SOTA) method in continual DAP scenario. However, in our experiments, DAS does not achieve SOTA performance due to the extensive evaluation of robustness across six domain orders.

In the end-task fine-tuning stage, Table~\ref{tab:numparam} shows the number of trainable parameters and the corresponding memory usage. DoMIX (last row) requires less than half the trainable parameters compared to LoRA ($r=48$) tuning, which has the same LoRA size as ours, due to the freezing of the $A$ module. Furthermore, DoMIX requires only 3.3\% of the trainable parameters compared to full fine-tuning, resulting in a 37\% reduction in peak GPU memory usage. While it may be a concern that the number of parameters of DoMIX increases linearly with the number of utilized domains, Table~\ref{tab:numparam} demonstrates that this results in only a small overhead in memory usage for each added domain module. Further explanations are provided in Appendix~\ref{section: Appendix: numDomainGPU}.


\noindent\textbf{Robustness.} As shown in the left two plots of Figure~\ref{fig:variancegraph}, DoMIX is not affected by variations in the domain sequence. In these plots, the error bars indicate the standard deviation computed over six values—each representing the average performance across six end tasks under a specific domain order. Specifically, for each of six randomly sampled domain orders, we perform continual domain-adaptive pre-training, then fine-tune the resulting model on all six end tasks. We then compute a single average performance score per domain order, resulting in six such scores in total. The standard deviation across these six average scores is visualized using error bars. As illustrated, all continual methods exhibit large variance across domain orders, indicating their sensitivity to the order in which data are presented. This characteristic is critical in real-world scenarios, where the sequence of data cannot be predetermined. In contrast, DoMIX demonstrates robust performance regardless of the domain sequence, as it does not rely on previously seen domains.

\begin{table*}[h!]
\caption{Accuracy comparison on commonsense reasoning tasks using LLaMA3-8B and Gemma2-9B. "Mem." denotes peak GPU memory usage during training. Bold numbers indicate the best performance for each task.}
\centering
\resizebox{\textwidth}{!}{
\begin{tabular}{l|l|c c c|c c c c c c c c|c}
\toprule
\textbf{Model} & \textbf{Method} & \textbf{Rank} & \textbf{Mem.} & \textbf{Time} & \textbf{WinoG.} & \textbf{OBQA} & \textbf{SIQA} & \textbf{ARC-e} & \textbf{BoolQ} & \textbf{ARC-c} & \textbf{PIQA} & \textbf{HellaS.} & \textbf{Avg.} \\
\midrule
\multirow{13}{*}{LLaMA3-8B} 
& AdapterP & -- & 43.4G & 7.0h & 81.69 & 81.60 & 78.10 & 82.03 & 68.38 & 69.97 & 83.08 & 90.48 & 79.41 \\
& AdapterH & -- & 48.3G & 8.0h & 75.77 & 73.80 & 76.15 & 73.19 & 65.44 & 58.96 & 80.25 & 76.69 & 72.53 \\
& Parallel & -- & 48.4G & 7.7h & 80.74 & 75.40 & 78.25 & 79.04 & 67.65 & 64.93 & 82.97 & 88.71 & 77.21 \\
& LoRA & 16 & 54.5G & 8.7h & 84.61 & 83.60 & 79.58 & 87.29 & 71.16 & 75.09 & 86.18 & 93.97 & 82.68 \\
& LoRA-Dash & 16 & 74.8G & 13.4h & \textbf{87.37} & \textbf{86.00} & 79.17 & 89.52 & 72.72 & 76.96 & 87.21 & 94.65 & 84.20 \\
& MoSLoRA & 16 & 54.5G & 8.7h & 86.19 & 84.40 & 80.50 & 89.56 & 68.56 & 79.95 & 88.36 & 95.25 & 84.10 \\
& DoRA & 16 & 66.5G & 14.3h & 85.64 & 85.80 & 80.30 & 90.66 & \textbf{74.95} & 79.52 & \textbf{89.34} & 95.52 & \textbf{85.22} \\
& LoRA (from Math LoRA) & 16 & 54.5G & 9.2h & 84.69 & 85.00 & 79.84 & 86.32 & 72.05 & 75.17 & 86.56 & 93.84 & 82.93 \\
& LoRA (with Joint Dataset) & 16 & 54.5G & 9.2h & 85.40 & 84.00 & \textbf{80.99} & 86.78 & 64.43 & 74.66 & 87.05 & 94.02 & 82.05 \\
& DoMIX (+ Math LoRA) & 16 & 54.5G & 9.2h & 85.95 & 85.80 & 80.25 & \textbf{91.04} & 73.94 & \textbf{80.55} & 88.36 & \textbf{95.85} & \textbf{85.22} \\
\cmidrule{2-14}
& LoRA & 32 & 54.6G & 8.7h & 82.95 & 81.00 & 79.53 & 85.10 & 69.85 & 70.73 & 85.85 & 92.00 & 80.88 \\
& DoRA & 32 & 66.6G & 14.3h & \textbf{87.92} & \textbf{86.60} & \textbf{80.96} & 90.45 & 73.73 & 77.82 & 88.74 & 95.63 & 85.23 \\
& DoMIX (+ Math LoRA) & 32 & 54.6G & 9.2h & 86.35 & 85.20 & 79.63 & \textbf{91.08} & \textbf{74.77} & \textbf{80.72} & \textbf{89.12} & \textbf{95.84} & \textbf{85.34} \\
\midrule
\multirow{4}{*}{Gemma2-9B} 
& LoRA & 16 & 51.8G & 13.5h & 87.69 & 90.00 & 81.78 & 92.89 & 75.44 & 84.30 & 89.72 & 95.58 & 87.17 \\
& DoRA & 16 & 60.3G & 21.3h & 88.24 & 91.20 & 82.14 & 95.20 & \textbf{78.13} & 87.71 & 90.48 & 96.62 & 88.72 \\
& LoRA (with Joint Dataset) & 16 & 51.8G & 14.3h & 89.11 & 88.20 & 81.42 & 93.10 & 75.72 & 82.76 & 90.21 & 95.31 & 86.98 \\
& DoMIX (+ Math LoRA) & 16 & 51.8G & 14.3h & \textbf{90.37} & \textbf{92.80} & \textbf{82.40} & \textbf{96.00} & 77.61 & \textbf{88.74} & \textbf{91.51} & \textbf{96.93} & \textbf{89.55} \\
\bottomrule
\end{tabular}
}
\label{tab:llmresult}
\end{table*}
\subsection{Extension to Standard LLM Fine-tuning}

In this section, we demonstrate that our method can be easily extended to standard LLM fine-tuning. We illustrate this on a common-sense reasoning task using LLaMA3-8B \citep{llama3modelcard}, following the setup of \citet{hu2023llmadapter}.
To further validate the scalability of DoMIX across different architectures and model sizes, we also conduct experiments with Gemma2-9B \citep{team2024gemma}. Common-sense reasoning involves asking LLMs to answer questions based on general world knowledge. This particular task consists of 8 subtasks, and the training set is constructed  from these 8 subtasks: \textit{Common-sense170K} \citep{hu2023llmadapter}. After fine-tuning the model on the training set, we evaluate its performance on each of the 8 individual tasks. We use a batch size of 16 for LLaMA3-8B and 8 for Gemma2-9B, and set the sequence length to 256 for all experiments.
Further experimental details are provided in Appendix~\ref{section:appendix:hyperparams reasoning}.


\subsubsection{Baselines}

We primarily compare our method against \textbf{LoRA} and its variants. \textbf{DoRA} \citep{liu2024dora} achieves state-of-the-art results in this setting by decomposing weights into magnitude and direction. \textbf{LoRA-Dash} \citep{si2024unleashing} identifies task-specific directions in an initial stage to enhance performance. \textbf{MoSLoRA} \citep{wu2024moslora} introduces a mixer matrix to utilize multiple subspaces during fine-tuning. Although MoSLoRA is closely related to our method, our approach differs by inserting a bridge matrix to appropriately exploit domain knowledge. We employ an intuitive initialization strategy, rather than random initialization, and adopt a different training procedure by freezing part of the modules. In addition to LoRA-based methods, we also compare our approach with non-LoRA PEFT techniques. \textbf{AdapterH} \citep{AdapterH} inserts a fully connected (FC) layer after both the attention and FFN layers, while \textbf{AdapterP} \citep{pfeiffer2020madadpaterP} inserts an FC layer only after the self-attention layer. \textbf{Parallel Adapter} \citep{he2021towardsParallelAdapter} introduces parallel learnable modules into the layers of the backbone model.

\subsubsection{Details of the Fine-tuning Process}
We use \textit{Common-sense170K} as one domain dataset and \textit{Math10k} \citep{hu2023llmadapter}, which contains mathematical questions and solutions, as another domain dataset. Following \citet{hu2023llmadapter}, baseline models are trained for three epochs on \textit{Common-sense170K}. For DoMIX, the model is first trained for two epochs using LoRA, after which the module trained on \textit{Math10k} is incorporated to apply our method during the final epoch. This ensures that DoMIX undergoes the same number of training iterations on \textit{Common-sense170K} as the baselines, allowing for a fair comparison.

\subsubsection{Results}
DoMIX either matches the state-of-the-art or achieves the best average performance across all settings. Specifically, it matches DoRA, the state-of-the-art method, at rank 16 on LLaMA3-8B, and outperforms all baselines at rank 32 on LLaMA3-8B and rank 16 on Gemma2-9B. This result is notable given that DoMIX requires substantially fewer computational resources—reducing GPU memory usage by 18\% and training time by 36\%—compared to DoRA \citep{liu2024dora}. The training time of DoMIX is the sum of the training time for LoRA on \textit{Math10k} and the training time for DoMIX on \textit{Common-sense170k}, which amounts to 0.5h + 8.7h for LLaMA3-8B and 0.8h + 13.5h for Gemma2-9B.

We also compared DoMIX with two cases: training from a LoRA module pre-trained on the \textit{Math10k} dataset, referred to as \textbf{LoRA (from Math LoRA)}, and training a LoRA module on a joint dataset (\textit{Math10k} and \textit{Common-sense170k}), referred to as \textbf{LoRA (with Joint Dataset)}. DoMIX demonstrates significantly better performance (+2.29\%p and +3.17\%p, respectively), for LLaMA3-8B at rank 16, suggesting that DoMIX is an effective framework for exploiting domain knowledge.

\begin{table}[ht]
    \centering
    \begin{minipage}{0.49\columnwidth}
        \centering
\caption{Average accuracy and F1 scores based on the presence and trainability of $P$.}
        \label{tab:ablation_P}
        \resizebox{\columnwidth}{!}{
        \begin{tabular}{lcS[table-format=2.2]S[table-format=2.2]}
            \toprule
            \textbf{$P$ init.} & \textbf{Trainable} & \textbf{Acc (\%)} & \textbf{F1 (\%)} \\
            \midrule
            \multirow{2}{*}{Id.} & \xmark & 71.47 & 61.99 \\
                                      & \cmark & 73.61 & 65.44 \\
            \midrule
            \multirow{2}{*}{Ours}     & \xmark & 69.59 & 52.57 \\
                                      & \cmark & \textbf{81.67} & \textbf{77.84} \\
            \bottomrule
        \end{tabular}
        }
    \end{minipage}%
    \hfill
    \begin{minipage}{0.49\columnwidth}
        \centering
        \caption{Performance comparison of trainable configurations.}
        \label{tab:ablation_freezing}
        \resizebox{\columnwidth}{!}{
        \begin{tabular}{cccSS}
            \toprule
            \multicolumn{3}{c}{\textbf{Trainable}} & \multicolumn{2}{c}{\textbf{Average (\%)}} \\
            \cmidrule(lr){1-3} \cmidrule(lr){4-5}
            \textbf{A} & \textbf{P} & \textbf{B} & \multicolumn{1}{c}{\textbf{Acc}} & \multicolumn{1}{c}{\textbf{F1}} \\
            \midrule
            \cmark & \xmark & \cmark & 81.39 & 77.63 \\
            \cmark & \cmark & \xmark & 80.74 & 76.57 \\
            \cmark & \cmark & \cmark & 81.31 & 77.00 \\
            \xmark & \cmark & \cmark & \textbf{81.67} & \textbf{77.84} \\
            \bottomrule
        \end{tabular}
        }
    \end{minipage}
\end{table}

\begin{table}[ht]
    \centering
    \caption{Performance comparison of different initialization configurations.}
    \label{tab:ablation_initialization}
    \resizebox{0.8\columnwidth}{!}{
    \begin{tabular}{cccS[table-format=2.2]S[table-format=2.2]}
        \toprule
        \multicolumn{3}{c}{\textbf{Initialization}} & \multicolumn{2}{c}{\textbf{Average (\%)}} \\
        \cmidrule(lr){1-3}\cmidrule(lr){4-5}
        \textbf{A} & \textbf{P} & \textbf{B} & \multicolumn{1}{c}{\textbf{Acc}} & \multicolumn{1}{c}{\textbf{F1}} \\
        \midrule
        Ours & Zero & Ours    & 81.27 & 77.30 \\
        Ours & Kaiming & Ours & 80.10 & 75.58 \\
        Ours & Ours & Zero    & 81.13 & 77.04 \\
        Ours & Ours & Kaiming & 81.11 & 77.12 \\
        Kaiming & Ours & Ours & 81.15 & 77.29 \\
        Ours & Ours & Ours    & \textbf{81.67} & \textbf{77.84} \\
        \bottomrule
    \end{tabular}
    }
\end{table}


\subsection{Ablation Studies}
\subsubsection{Effect of the Bridge Module $P$}
\label{section:ablation:effectofP}

In this section, we analyze the effect of the bridge module $P$ by varying its presence and trainability. In Table~\ref{tab:ablation_P}, the first row represents the absence of a bridge module—specifically, $P$ is initialized as the identity matrix and kept frozen. Comparing this configuration with our proposed method (last row) clearly demonstrates the benefit of introducing a bridge module. Furthermore, the performance difference between the first and second rows (identity initialization, with and without training), as well as between the third and fourth rows (our initialization, with and without training), indicates that the trainability of $P$ is crucial. Lastly, a comparison between the second and fourth rows—where the only difference is the initialization of trainable $P$—highlights the importance of initialization. A more detailed analysis on initialization strategies is provided in Section~\ref{section:ablation:effect of Initialization}.

In summary, these results validate both the necessity of the bridge module $P$ and the importance of making it trainable.

\subsubsection{Effect of Freezing}
In this section, we verify the rationale for freezing only the \(A\) modules (those closer to the input). Table~\ref{tab:ablation_freezing} presents various freezing configurations. Freezing the bridge module \(P\) (first row) or the \(B\) modules (second row) results in inferior performance compared to freezing the \(A\) modules (our method, fourth row). These findings suggest that the performance gains of our method are not merely a result of freezing any module. Moreover, our approach (fourth row) achieves superior performance compared to the configuration where the \(A\) modules remain trainable (third row). This validates our intuition that fine-tuning within the pre-trained knowledge subspace—enabled by freezing the \(A\) modules—facilitates more effective domain knowledge exploitation and leads to improved fine-tuning performance.

\subsubsection{Effect of Initialization}
\label{section:ablation:effect of Initialization}
In this section, we evaluate the effectiveness of our initialization strategy, as shown in Table~\ref{tab:ablation_initialization}. We tested three initialization methods—Ours, Zero, and Kaiming—for each module (\(A\), \(P\), and \(B\)). In the "Ours" initialization, $A$ and $B$ are initialized by concatenating domain-specific pre-trained weights, while $P$ is initialized as a trainable diagonal matrix whose equal-valued diagonal entries sum to 1. "Zero" represents zeroed initialization, and "Kaiming" corresponds to Kaiming uniform initialization \citep{he2015delvingkaiminginit}. 

By comparing the first, second, and last rows, we observe that our initialization of $P$ is effective. This result differs from \citet{wu2024moslora}, where random initialization was found to be optimal. The discrepancy arises because their method, MoSLoRA, operates in standard fine-tuning settings without leveraging pre-trained domain-specific modules. In contrast, our framework is designed to exploit pre-trained domain knowledge, for which a diagonal initialization of \(P\) is more appropriate and effective. 

Furthermore, by comparing the third, fourth, and fifth rows with the last row, we observe that random or zeroed initialization of the \(A\) and \(B\) modules results in inferior performance compared to our method. These results confirm that our performance gains stem from effectively exploiting pre-trained knowledge using an appropriate initialization of the bridge matrix.


\subsection{Analysis on the Bridge Module $P$}
\begin{figure}[h!]
    \centering
    \begin{subfigure}[b]{0.99\columnwidth}
        \centering
        \includegraphics[width=\columnwidth]{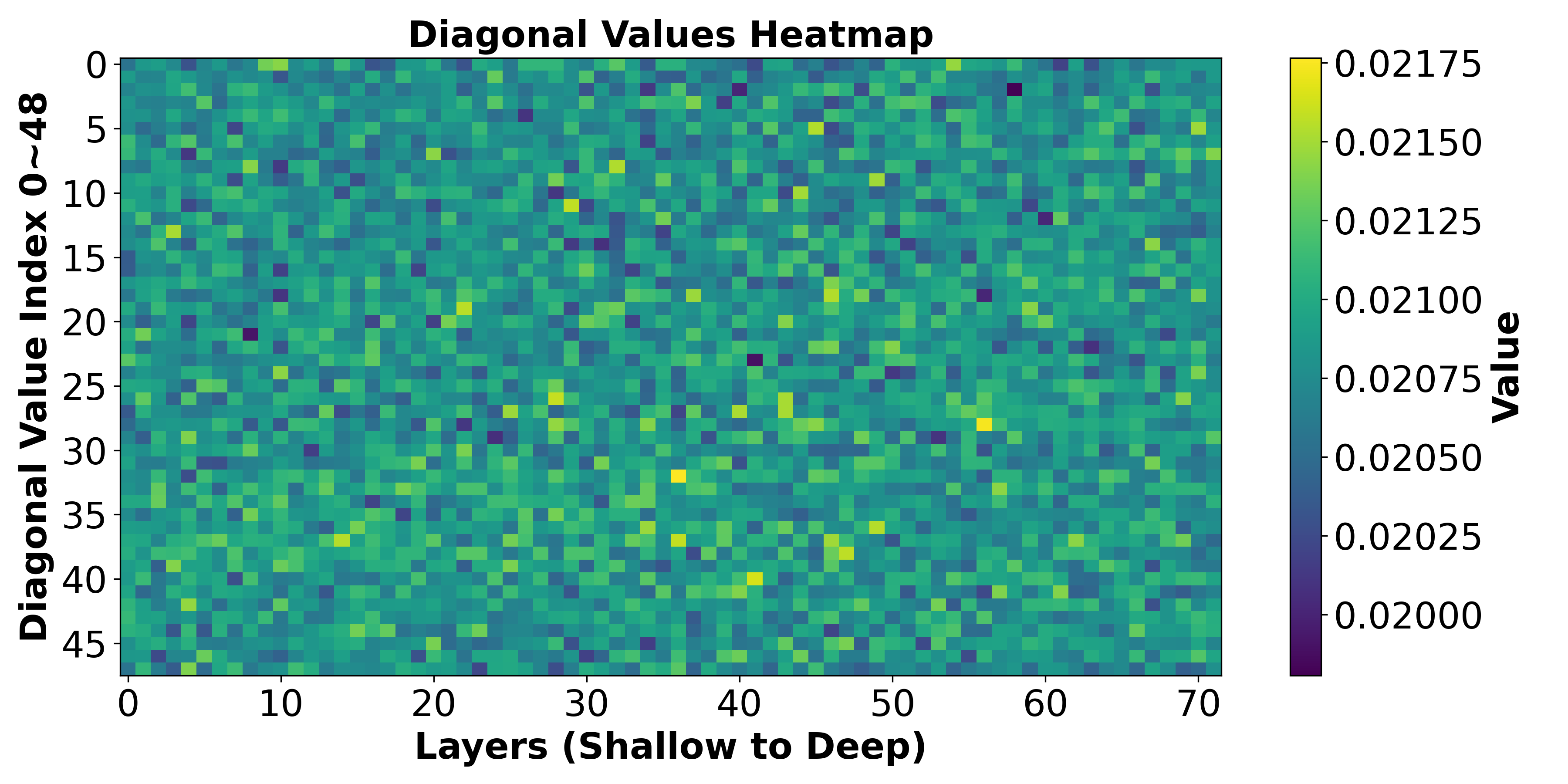}
        \label{fig:DAP_diag}
    \end{subfigure}

    \vspace{-0.2cm} 

    \begin{subfigure}[b]{0.99\columnwidth}
        \centering
        \includegraphics[width=\columnwidth]{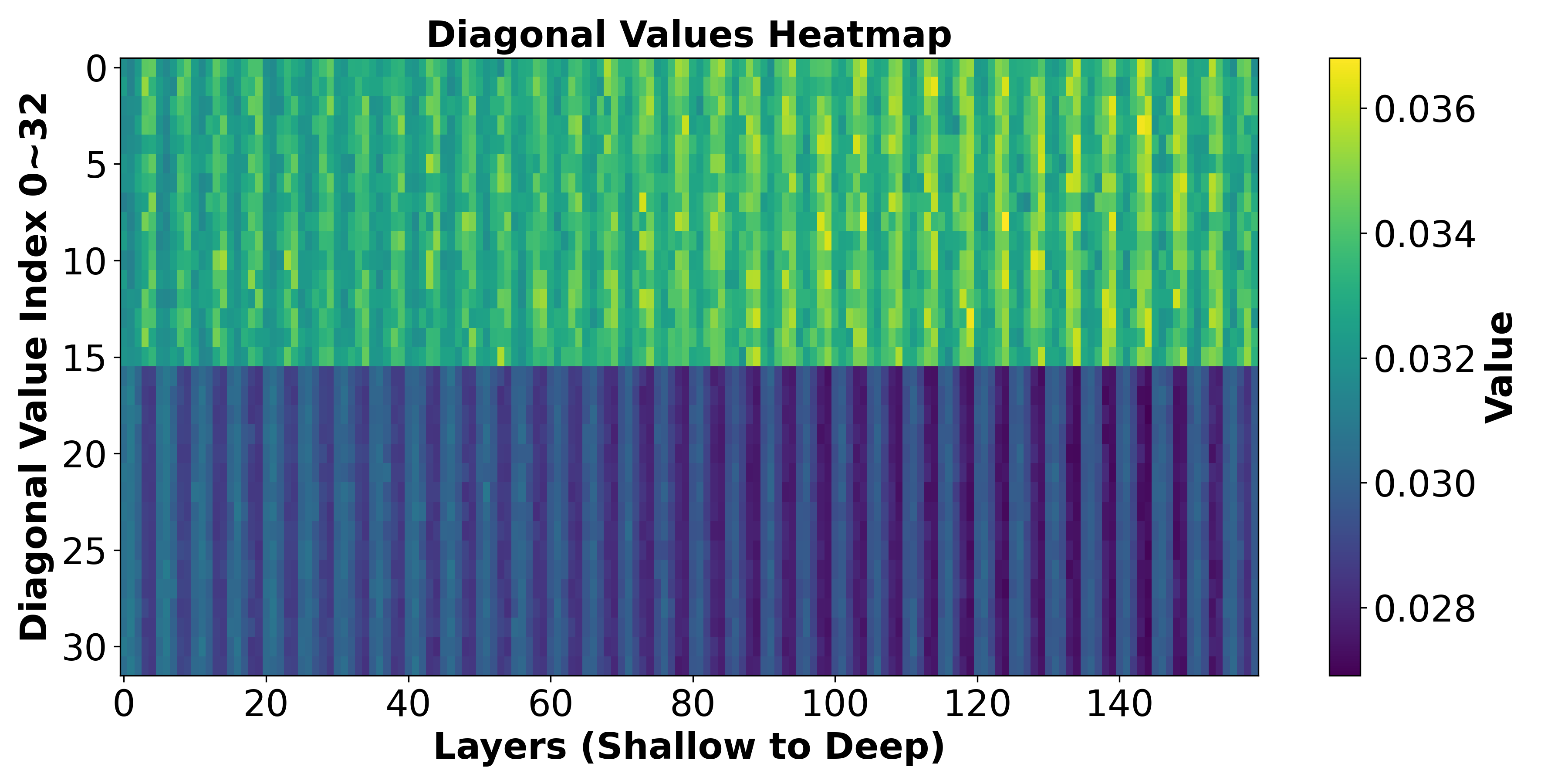}
        \label{fig:LLM_diag_heatmap}
    \end{subfigure}

    \vspace{-0.5cm}
    \caption{Comparison of diagonal values in $P$ heatmaps: DAP setting (top) and LLM fine-tuning setting (bottom).}
    \label{fig:diag_heatmaps}
\end{figure}


\noindent Since the bridge module \(P\) is introduced to control the extent of domain knowledge exploitation, we conducted an analysis of its values. Figure~\ref{fig:diag_heatmaps} presents the diagonal values of \(P\), where we apply softmax to the diagonal entries within each layer. The y-axis corresponds to the diagonal indices of the \(P\) matrix. These values indicate how strongly each subspace (i.e., domain-specific knowledge) is emphasized during fine-tuning.

In the DAP setting (upper plot), no clear patterns were observed in the diagonal values of \(P\), indicating that various domain knowledge was utilized. However, in the LLM fine-tuning setting (lower plot), a distinct pattern emerged: the diagonal values corresponding to the LoRA module trained on \textit{Common-sense170k} were generally higher than those corresponding to the LoRA module trained on \textit{Math10k}. While the difference is noticeable, the values for the \textit{Math10k} module were still non-trivial, considering that all diagonal values ranged approximately from 0.028 to 0.036. This indicates that the model incorporated mathematical knowledge to some extent.

These results suggest that the model effectively exploited the appropriate domain knowledge during fine-tuning without requiring any prior information. Furthermore, the presence of non-zero values for Math LoRA indicates that DoMIX also leveraged this knowledge, contributing to its improved performance.

\section{Conclusion}
In this paper, we emphasize the need for an efficient, parallel, and effective methodology for domain learning and knowledge exploitation. To address this, we proposed DoMIX, which achieves performance on par with or better than state-of-the-art methods, while demonstrating superior efficiency in terms of training time and GPU memory usage. We validated the efficacy of our method in both the continual DAP setting and the standard LLM fine-tuning setting.

We hope this work contributes to advancing research on enabling models to generalize effectively across diverse tasks by leveraging pre-trained models, which are abundantly available online, as inspired by the notion of “\textit{standing on the shoulders of giants}.”


\section{Limitations}

While DoMIX demonstrates efficiency in terms of memory usage and training time, both in continual DAP settings and standard LLM fine-tuning, it has some minor limitations. First, it requires linearly increasing parameters to accommodate additional domain knowledge, although the overhead in parameter and memory cost per domain is relatively small. Second, the pre-trained LoRA modules must be stored for use during the fine-tuning step. 

To address these limitations, a promising direction for future work could involve identifying important subspaces to retain and dismissing redundant subspaces through theoretical analysis. Such an approach could reduce the number of parameters required for each domain and further lower the memory cost during fine-tuning.

\section*{Acknowledgments}
This work was supported in part by the National Research Foundation of Korea (NRF) grant [No.2021R1A2C2007884], the Institute of Information \& communications Technology Planning \& Evaluation (IITP) grants [RS-2021-II211343, RS-2021-II212068, RS-2022-II220113, RS-2022-II220959], and the BK21 FOUR Education and Research Program for Future
ICT Pioneers (Seoul National University), funded by the Korean government (MSIT). It was also supported by Mobile eXperience (MX) Business, Samsung Electronics Co., Ltd. 
\bibliography{custom}

\begin{thebibliography}{47}
\providecommand{\natexlab}[1]{#1}

\bibitem[{Achiam et~al.(2023)Achiam, Adler, Agarwal, Ahmad, Akkaya, Aleman, Almeida, Altenschmidt, Altman, Anadkat et~al.}]{achiam2023gpt}
Josh Achiam, Steven Adler, Sandhini Agarwal, Lama Ahmad, Ilge Akkaya, Florencia~Leoni Aleman, Diogo Almeida, Janko Altenschmidt, Sam Altman, Shyamal Anadkat, et~al. 2023.
\newblock \href {https://arxiv.org/abs/2303.08774} {{GPT}-4 technical report}.
\newblock \emph{arXiv preprint arXiv:2303.08774}.

\bibitem[{AI@Meta(2024)}]{llama3modelcard}
AI@Meta. 2024.
\newblock \href {https://github.com/meta-llama/llama3/blob/main/MODEL_CARD.md} {Llama 3 model card}.

\bibitem[{Alsentzer et~al.(2019)Alsentzer, Murphy, Boag, Weng, Jindi, Naumann, and McDermott}]{alsentzer2019publicly}
Emily Alsentzer, John Murphy, William Boag, Wei-Hung Weng, Di~Jindi, Tristan Naumann, and Matthew McDermott. 2019.
\newblock \href {https://doi.org/10.18653/v1/W19-1909} {Publicly available clinical {BERT} embeddings}.
\newblock In \emph{Proceedings of the 2nd Clinical Natural Language Processing Workshop}, pages 72--78, Minneapolis, Minnesota, USA. Association for Computational Linguistics.

\bibitem[{Beltagy et~al.(2019)Beltagy, Lo, and Cohan}]{beltagy2019scibertf1}
Iz~Beltagy, Kyle Lo, and Arman Cohan. 2019.
\newblock \href {https://doi.org/10.18653/v1/D19-1371} {{S}ci{BERT}: A pretrained language model for scientific text}.
\newblock In \emph{{Proceedings of the 2019 Conference on Empirical Methods in Natural Language Processing and the 9th International Joint Conference on Natural Language Processing (EMNLP-IJCNLP)}}, pages 3615--3620, Hong Kong, China. Association for Computational Linguistics.

\bibitem[{Brown et~al.(2020)Brown, Mann, Ryder, Subbiah, Kaplan, Dhariwal, Neelakantan, Shyam, Sastry, Askell, Agarwal, Herbert-Voss, Krueger, Henighan, Child, Ramesh, Ziegler, Wu, Winter, Hesse, Chen, Sigler, Litwin, Gray, Chess, Clark, Berner, McCandlish, Radford, Sutskever, and Amodei}]{gpt3}
Tom Brown, Benjamin Mann, Nick Ryder, Melanie Subbiah, Jared~D Kaplan, Prafulla Dhariwal, Arvind Neelakantan, Pranav Shyam, Girish Sastry, Amanda Askell, Sandhini Agarwal, Ariel Herbert-Voss, Gretchen Krueger, Tom Henighan, Rewon Child, Aditya Ramesh, Daniel Ziegler, Jeffrey Wu, Clemens Winter, Chris Hesse, Mark Chen, Eric Sigler, Mateusz Litwin, Scott Gray, Benjamin Chess, Jack Clark, Christopher Berner, Sam McCandlish, Alec Radford, Ilya Sutskever, and Dario Amodei. 2020.
\newblock \href {https://proceedings.neurips.cc/paper_files/paper/2020/file/1457c0d6bfcb4967418bfb8ac142f64a-Paper.pdf} {Language models are few-shot learners}.
\newblock In \emph{Advances in Neural Information Processing Systems}, volume~33, pages 1877--1901.

\bibitem[{Buzzega et~al.(2020)Buzzega, Boschini, Porrello, Abati, and Calderara}]{buzzega2020dark}
Pietro Buzzega, Matteo Boschini, Angelo Porrello, Davide Abati, and Simone Calderara. 2020.
\newblock \href {https://proceedings.neurips.cc/paper/2020/hash/b704ea2c39778f07c617f6b7ce480e9e-Abstract.html} {Dark experience for general continual learning: a strong, simple baseline}.
\newblock In \emph{Advances in Neural Information Processing Systems}, volume~33, pages 15920--15930.

\bibitem[{Dery et~al.(2022)Dery, Michel, Talwalkar, and Neubig}]{dery2021shouldf1}
Lucio~M. Dery, Paul Michel, Ameet Talwalkar, and Graham Neubig. 2022.
\newblock \href {https://openreview.net/forum?id=2bO2x8NAIMB} {Should we be pre-training? an argument for end-task aware training as an alternative}.
\newblock In \emph{International Conference on Learning Representations}.

\bibitem[{Devlin et~al.(2019)Devlin, Chang, Lee, and Toutanova}]{devlin-etal-2019-bert}
Jacob Devlin, Ming-Wei Chang, Kenton Lee, and Kristina Toutanova. 2019.
\newblock \href {https://doi.org/10.18653/v1/N19-1423} {{BERT}: Pre-training of deep bidirectional transformers for language understanding}.
\newblock In \emph{Proceedings of the 2019 Conference of the North {A}merican Chapter of the Association for Computational Linguistics: Human Language Technologies, Volume 1 (Long and Short Papers)}, pages 4171--4186, Minneapolis, Minnesota. Association for Computational Linguistics.

\bibitem[{Ding et~al.(2008)Ding, Liu, and Yu}]{ding2008holistic}
Xiaowen Ding, Bing Liu, and Philip~S Yu. 2008.
\newblock \href {https://www.cs.uic.edu/~liub/FBS/opinion-mining-final-WSDM.pdf} {A holistic lexicon-based approach to opinion mining}.
\newblock In \emph{Proceedings of the 2008 international conference on web search and data mining}, pages 231--240.

\bibitem[{Gururangan et~al.(2020)Gururangan, Marasovi{\'c}, Swayamdipta, Lo, Beltagy, Downey, and Smith}]{gururangan2020don}
Suchin Gururangan, Ana Marasovi{\'c}, Swabha Swayamdipta, Kyle Lo, Iz~Beltagy, Doug Downey, and Noah~A. Smith. 2020.
\newblock \href {https://doi.org/10.18653/v1/2020.acl-main.740} {Don{'}t stop pretraining: Adapt language models to domains and tasks}.
\newblock In \emph{Proceedings of the 58th Annual Meeting of the Association for Computational Linguistics}, pages 8342--8360, Online. Association for Computational Linguistics.

\bibitem[{He et~al.(2022)He, Zhou, Ma, Berg-Kirkpatrick, and Neubig}]{he2021towardsParallelAdapter}
Junxian He, Chunting Zhou, Xuezhe Ma, Taylor Berg-Kirkpatrick, and Graham Neubig. 2022.
\newblock \href {https://openreview.net/forum?id=0RDcd5Axok} {Towards a unified view of parameter-efficient transfer learning}.
\newblock In \emph{International Conference on Learning Representations}.

\bibitem[{He et~al.(2015)He, Zhang, Ren, and Sun}]{he2015delvingkaiminginit}
Kaiming He, Xiangyu Zhang, Shaoqing Ren, and Jian Sun. 2015.
\newblock \href {https://www.cv-foundation.org/openaccess/content_iccv_2015/papers/He_Delving_Deep_into_ICCV_2015_paper.pdf} {Delving deep into rectifiers: Surpassing human-level performance on imagenet classification}.
\newblock In \emph{Proceedings of the IEEE international conference on computer vision}, pages 1026--1034.

\bibitem[{Hinton(2015)}]{hinton2015distilling}
Geoffrey Hinton. 2015.
\newblock \href {https://arxiv.org/abs/1503.02531} {Distilling the knowledge in a neural network}.
\newblock \emph{arXiv preprint arXiv:1503.02531}.

\bibitem[{Houlsby et~al.(2019)Houlsby, Giurgiu, Jastrzebski, Morrone, De~Laroussilhe, Gesmundo, Attariyan, and Gelly}]{AdapterH}
Neil Houlsby, Andrei Giurgiu, Stanislaw Jastrzebski, Bruna Morrone, Quentin De~Laroussilhe, Andrea Gesmundo, Mona Attariyan, and Sylvain Gelly. 2019.
\newblock \href {https://proceedings.mlr.press/v97/houlsby19a.html} {Parameter-efficient transfer learning for {NLP}}.
\newblock In \emph{International Conference on Machine Learning}.

\bibitem[{Hu et~al.(2022)Hu, Shen, Wallis, Allen-Zhu, Li, Wang, Wang, and Chen}]{hu2021lora}
Edward~J Hu, Yelong Shen, Phillip Wallis, Zeyuan Allen-Zhu, Yuanzhi Li, Shean Wang, Lu~Wang, and Weizhu Chen. 2022.
\newblock \href {https://openreview.net/forum?id=nZeVKeeFYf9} {Lo{RA}: Low-rank adaptation of large language models}.
\newblock In \emph{International Conference on Learning Representations}.

\bibitem[{Hu and Liu(2004)}]{10.1145/1014052.1014073}
Minqing Hu and Bing Liu. 2004.
\newblock \href {https://doi.org/10.1145/1014052.1014073} {Mining and summarizing customer reviews}.
\newblock In \emph{Proceedings of the Tenth ACM SIGKDD International Conference on Knowledge Discovery and Data Mining}, KDD '04, page 168–177, New York, NY, USA. Association for Computing Machinery.

\bibitem[{Hu et~al.(2023)Hu, Wang, Lan, Xu, Lim, Bing, Xu, Poria, and Lee}]{hu2023llmadapter}
Zhiqiang Hu, Lei Wang, Yihuai Lan, Wanyu Xu, Ee-Peng Lim, Lidong Bing, Xing Xu, Soujanya Poria, and Roy Lee. 2023.
\newblock \href {https://doi.org/10.18653/v1/2023.emnlp-main.319} {{LLM}-adapters: An adapter family for parameter-efficient fine-tuning of large language models}.
\newblock In \emph{Proceedings of the 2023 Conference on Empirical Methods in Natural Language Processing}, pages 5254--5276, Singapore. Association for Computational Linguistics.

\bibitem[{Jurgens et~al.(2018)Jurgens, Kumar, Hoover, McFarland, and Jurafsky}]{jurgens-etal-2018-measuring}
David Jurgens, Srijan Kumar, Raine Hoover, Dan McFarland, and Dan Jurafsky. 2018.
\newblock \href {https://doi.org/10.1162/tacl_a_00028} {Measuring the evolution of a scientific field through citation frames}.
\newblock \emph{Transactions of the Association for Computational Linguistics}, 6:391--406.

\bibitem[{Ke et~al.(2022{\natexlab{a}})Ke, Lin, Shao, Xu, Shu, and Liu}]{CPT}
Zixuan Ke, Haowei Lin, Yijia Shao, Hu~Xu, Lei Shu, and Bing Liu. 2022{\natexlab{a}}.
\newblock \href {https://doi.org/10.18653/v1/2022.emnlp-main.695} {Continual training of language models for few-shot learning}.
\newblock In \emph{Proceedings of the 2022 Conference on Empirical Methods in Natural Language Processing}, pages 10205--10216, Abu Dhabi, United Arab Emirates. Association for Computational Linguistics.

\bibitem[{Ke et~al.(2021{\natexlab{a}})Ke, Liu, Ma, Xu, and Shu}]{CTR}
Zixuan Ke, Bing Liu, Nianzu Ma, Hu~Xu, and Lei Shu. 2021{\natexlab{a}}.
\newblock \href {https://openreview.net/forum?id=RJ7XFI15Q8f} {Achieving forgetting prevention and knowledge transfer in continual learning}.
\newblock In \emph{Advances in Neural Information Processing Systems}, volume~34, pages 22443--22456.

\bibitem[{Ke et~al.(2021{\natexlab{b}})Ke, Liu, Xu, and Shu}]{ke2021classic}
Zixuan Ke, Bing Liu, Hu~Xu, and Lei Shu. 2021{\natexlab{b}}.
\newblock \href {https://doi.org/10.18653/v1/2021.emnlp-main.550} {{CLASSIC}: Continual and contrastive learning of aspect sentiment classification tasks}.
\newblock In \emph{Proceedings of the 2021 Conference on Empirical Methods in Natural Language Processing}, pages 6871--6883, Online and Punta Cana, Dominican Republic. Association for Computational Linguistics.

\bibitem[{Ke et~al.(2023)Ke, Shao, Lin, Konishi, Kim, and Liu}]{ke2023continual}
Zixuan Ke, Yijia Shao, Haowei Lin, Tatsuya Konishi, Gyuhak Kim, and Bing Liu. 2023.
\newblock \href {https://openreview.net/forum?id=m_GDIItaI3o} {Continual pre-training of language models}.
\newblock In \emph{International Conference on Learning Representations}.

\bibitem[{Ke et~al.(2022{\natexlab{b}})Ke, Shao, Lin, Xu, Shu, and Liu}]{DGA}
Zixuan Ke, Yijia Shao, Haowei Lin, Hu~Xu, Lei Shu, and Bing Liu. 2022{\natexlab{b}}.
\newblock \href {https://doi.org/10.18653/v1/2022.emnlp-main.693} {Adapting a language model while preserving its general knowledge}.
\newblock In \emph{Proceedings of the 2022 Conference on Empirical Methods in Natural Language Processing}, pages 10177--10188, Abu Dhabi, United Arab Emirates. Association for Computational Linguistics.

\bibitem[{Ke et~al.(2021{\natexlab{c}})Ke, Xu, and Liu}]{keBCL}
Zixuan Ke, Hu~Xu, and Bing Liu. 2021{\natexlab{c}}.
\newblock \href {https://doi.org/10.18653/v1/2021.naacl-main.378} {Adapting {BERT} for continual learning of a sequence of aspect sentiment classification tasks}.
\newblock In \emph{Proceedings of the 2021 Conference of the North American Chapter of the Association for Computational Linguistics: Human Language Technologies}, pages 4746--4755, Online. Association for Computational Linguistics.

\bibitem[{Kirkpatrick et~al.(2017)Kirkpatrick, Pascanu, Rabinowitz, Veness, Desjardins, Rusu, Milan, Quan, Ramalho, Grabska-Barwinska, Hassabis, Clopath, Kumaran, and Hadsell}]{EWC}
James Kirkpatrick, Razvan Pascanu, Neil Rabinowitz, Joel Veness, Guillaume Desjardins, Andrei~A. Rusu, Kieran Milan, John Quan, Tiago Ramalho, Agnieszka Grabska-Barwinska, Demis Hassabis, Claudia Clopath, Dharshan Kumaran, and Raia Hadsell. 2017.
\newblock \href {https://doi.org/10.1073/pnas.1611835114} {Overcoming catastrophic forgetting in neural networks}.
\newblock \emph{Proceedings of the National Academy of Sciences}, 114(13):3521--3526.

\bibitem[{Kopiczko et~al.(2024)Kopiczko, Blankevoort, and Asano}]{kopiczko2024vera}
Dawid~Jan Kopiczko, Tijmen Blankevoort, and Yuki~M Asano. 2024.
\newblock \href {https://openreview.net/forum?id=NjNfLdxr3A} {Ve{RA}: Vector-based random matrix adaptation}.
\newblock In \emph{International Conference on Learning Representations}.

\bibitem[{Kringelum et~al.(2016)Kringelum, Kjaerulff, Brunak, Lund, Oprea, and Taboureau}]{kringelum2016chemprot}
Jens Kringelum, Sonny~Kim Kjaerulff, S{\o}ren Brunak, Ole Lund, Tudor~I Oprea, and Olivier Taboureau. 2016.
\newblock \href {https://pubmed.ncbi.nlm.nih.gov/26876982/} {{ChemProt-3.0}: a global chemical biology diseases mapping}.
\newblock \emph{Database}, 2016:bav123.

\bibitem[{Lee et~al.(2020)Lee, Yoon, Kim, Kim, Kim, So, and Kang}]{lee2020biobert}
Jinhyuk Lee, Wonjin Yoon, Sungdong Kim, Donghyeon Kim, Sunkyu Kim, Chan~Ho So, and Jaewoo Kang. 2020.
\newblock \href {https://academic.oup.com/bioinformatics/article/36/4/1234/5566506} {Bio{BERT}: a pre-trained biomedical language representation model for biomedical text mining}.
\newblock \emph{Bioinformatics}, 36(4):1234--1240.

\bibitem[{Lester et~al.(2021)Lester, Al-Rfou, and Constant}]{Prompttuning}
Brian Lester, Rami Al-Rfou, and Noah Constant. 2021.
\newblock \href {https://doi.org/10.18653/v1/2021.emnlp-main.243} {The power of scale for parameter-efficient prompt tuning}.
\newblock In \emph{Proceedings of the 2021 Conference on Empirical Methods in Natural Language Processing}, pages 3045--3059, Online and Punta Cana, Dominican Republic. Association for Computational Linguistics.

\bibitem[{Li and Liang(2021)}]{Prefixtuning}
Xiang~Lisa Li and Percy Liang. 2021.
\newblock \href {https://doi.org/10.18653/v1/2021.acl-long.353} {Prefix-tuning: Optimizing continuous prompts for generation}.
\newblock In \emph{Proceedings of the 59th Annual Meeting of the Association for Computational Linguistics and the 11th International Joint Conference on Natural Language Processing (Volume 1: Long Papers)}, pages 4582--4597, Online. Association for Computational Linguistics.

\bibitem[{Liang and Li(2024)}]{liang2024inflora}
Yan-Shuo Liang and Wu-Jun Li. 2024.
\newblock \href {https://openaccess.thecvf.com/content/CVPR2024/papers/Liang_InfLoRA_Interference-Free_Low-Rank_Adaptation_for_Continual_Learning_CVPR_2024_paper.pdf} {{InfLoRA}: Interference-free low-rank adaptation for continual learning}.
\newblock In \emph{Proceedings of the IEEE/CVF Conference on Computer Vision and Pattern Recognition}, pages 23638--23647.

\bibitem[{Liu et~al.(2024)Liu, Wang, Yin, Molchanov, Wang, Cheng, and Chen}]{liu2024dora}
Shih-Yang Liu, Chien-Yi Wang, Hongxu Yin, Pavlo Molchanov, Yu-Chiang~Frank Wang, Kwang-Ting Cheng, and Min-Hung Chen. 2024.
\newblock \href {https://arxiv.org/abs/2402.09353} {Do{RA}: Weight-decomposed low-rank adaptation}.
\newblock In \emph{International Conference on Machine Learning}.

\bibitem[{Liu et~al.(2019)Liu, Ott, Goyal, Du, Joshi, Chen, Levy, Lewis, Zettlemoyer, and Stoyanov}]{liu2019roberta}
Yinhan Liu, Myle Ott, Naman Goyal, Jingfei Du, Mandar Joshi, Danqi Chen, Omer Levy, Mike Lewis, Luke Zettlemoyer, and Veselin Stoyanov. 2019.
\newblock \href {https://arxiv.org/abs/1907.11692} {{RoBERTa}: A robustly optimized bert pretraining approach}.
\newblock \emph{arXiv preprint arXiv:1907.11692}.

\bibitem[{Lo et~al.(2020)Lo, Wang, Neumann, Kinney, and Weld}]{lo2019s2orc}
Kyle Lo, Lucy~Lu Wang, Mark Neumann, Rodney Kinney, and Daniel Weld. 2020.
\newblock \href {https://doi.org/10.18653/v1/2020.acl-main.447} {{S}2{ORC}: The semantic scholar open research corpus}.
\newblock In \emph{Proceedings of the 58th Annual Meeting of the Association for Computational Linguistics}, pages 4969--4983, Online. Association for Computational Linguistics.

\bibitem[{Luan et~al.(2018)Luan, He, Ostendorf, and Hajishirzi}]{luan2018multi}
Yi~Luan, Luheng He, Mari Ostendorf, and Hannaneh Hajishirzi. 2018.
\newblock \href {https://doi.org/10.18653/v1/D18-1360} {Multi-task identification of entities, relations, and coreference for scientific knowledge graph construction}.
\newblock In \emph{Proceedings of the 2018 Conference on Empirical Methods in Natural Language Processing}, pages 3219--3232, Brussels, Belgium. Association for Computational Linguistics.

\bibitem[{Ni et~al.(2019)Ni, Li, and McAuley}]{ni2019justifying}
Jianmo Ni, Jiacheng Li, and Julian McAuley. 2019.
\newblock \href {https://doi.org/10.18653/v1/D19-1018} {Justifying recommendations using distantly-labeled reviews and fine-grained aspects}.
\newblock In \emph{Proceedings of the 2019 Conference on Empirical Methods in Natural Language Processing and the 9th International Joint Conference on Natural Language Processing (EMNLP-IJCNLP)}, pages 188--197, Hong Kong, China. Association for Computational Linguistics.

\bibitem[{Pfeiffer et~al.(2020)Pfeiffer, Vuli{\'c}, Gurevych, and Ruder}]{pfeiffer2020madadpaterP}
Jonas Pfeiffer, Ivan Vuli{\'c}, Iryna Gurevych, and Sebastian Ruder. 2020.
\newblock \href {https://doi.org/10.18653/v1/2020.emnlp-main.617} {{MAD-X}: {A}n {A}dapter-{B}ased {F}ramework for {M}ulti-{T}ask {C}ross-{L}ingual {T}ransfer}.
\newblock In \emph{Proceedings of the 2020 Conference on Empirical Methods in Natural Language Processing (EMNLP)}, pages 7654--7673, Online. Association for Computational Linguistics.

\bibitem[{Sap et~al.(2020)Sap, Shwartz, Bosselut, Choi, and Roth}]{sap2020commonsense}
Maarten Sap, Vered Shwartz, Antoine Bosselut, Yejin Choi, and Dan Roth. 2020.
\newblock \href {https://doi.org/10.18653/v1/2020.acl-tutorials.7} {Commonsense reasoning for natural language processing}.
\newblock In \emph{Proceedings of the 58th Annual Meeting of the Association for Computational Linguistics: Tutorial Abstracts}, pages 27--33, Online. Association for Computational Linguistics.

\bibitem[{Serra et~al.(2018)Serra, Suris, Miron, and Karatzoglou}]{serra2018hat}
Joan Serra, Didac Suris, Marius Miron, and Alexandros Karatzoglou. 2018.
\newblock \href {https://arxiv.org/abs/1801.01423} {Overcoming catastrophic forgetting with hard attention to the task}.
\newblock In \emph{International Conference on Machine Learning}.

\bibitem[{Si et~al.(2025)Si, Shi, Zhang, Yang, Pfister, and Shen}]{si2024unleashing}
Chongjie Si, Zhiyi Shi, Shifan Zhang, Xiaokang Yang, Hanspeter Pfister, and Wei Shen. 2025.
\newblock \href {https://openreview.net/forum?id=RYrJqz44p4} {Unleashing the power of task-specific directions in parameter efficient fine-tuning}.
\newblock In \emph{International Conference on Learning Representations}.

\bibitem[{Sun et~al.(2019)Sun, Qiu, Xu, and Huang}]{sun2019fine}
Chi Sun, Xipeng Qiu, Yige Xu, and Xuanjing Huang. 2019.
\newblock \href {https://arxiv.org/abs/1905.05583} {{How to fine-tune BERT for text classification?}}
\newblock In \emph{Chinese computational linguistics: 18th China national conference, CCL 2019, Kunming, China, October 18--20, 2019, proceedings 18}, pages 194--206. Springer.

\bibitem[{Team et~al.(2024)Team, Mesnard, Hardin, Dadashi, Bhupatiraju, Pathak, Sifre, Rivi{\`e}re, Kale, Love et~al.}]{team2024gemma}
Gemma Team, Thomas Mesnard, Cassidy Hardin, Robert Dadashi, Surya Bhupatiraju, Shreya Pathak, Laurent Sifre, Morgane Rivi{\`e}re, Mihir~Sanjay Kale, Juliette Love, et~al. 2024.
\newblock \href {https://arxiv.org/abs/2403.08295} {Gemma: Open models based on gemini research and technology}.
\newblock \emph{arXiv preprint arXiv:2403.08295}.

\bibitem[{Touvron et~al.(2023)Touvron, Lavril, Izacard, Martinet, Lachaux, Lacroix, Rozi{\`e}re, Goyal, Hambro, Azhar et~al.}]{touvron2023llama}
Hugo Touvron, Thibaut Lavril, Gautier Izacard, Xavier Martinet, Marie-Anne Lachaux, Timoth{\'e}e Lacroix, Baptiste Rozi{\`e}re, Naman Goyal, Eric Hambro, Faisal Azhar, et~al. 2023.
\newblock \href {https://arxiv.org/abs/2302.13971} {{LLaMA: Open and efficient foundation language models}}.
\newblock \emph{arXiv preprint arXiv:2302.13971}.

\bibitem[{Wang et~al.(2018)Wang, Singh, Michael, Hill, Levy, and Bowman}]{wang2018glue}
Alex Wang, Amanpreet Singh, Julian Michael, Felix Hill, Omer Levy, and Samuel Bowman. 2018.
\newblock \href {https://doi.org/10.18653/v1/W18-5446} {{GLUE}: A multi-task benchmark and analysis platform for natural language understanding}.
\newblock In \emph{Proceedings of the 2018 {EMNLP} Workshop {B}lackbox{NLP}: Analyzing and Interpreting Neural Networks for {NLP}}, pages 353--355, Brussels, Belgium. Association for Computational Linguistics.

\bibitem[{Wei et~al.(2022)Wei, Wang, Schuurmans, Bosma, Xia, Chi, Le, Zhou et~al.}]{wei2022chain}
Jason Wei, Xuezhi Wang, Dale Schuurmans, Maarten Bosma, Fei Xia, Ed~Chi, Quoc~V Le, Denny Zhou, et~al. 2022.
\newblock \href {https://arxiv.org/abs/2201.11903} {Chain-of-thought prompting elicits reasoning in large language models}.
\newblock In \emph{Advances in Neural Information Processing Systems}, volume~35, pages 24824--24837.

\bibitem[{Wu et~al.(2024)Wu, Wang, Zhao, and Wong}]{wu2024moslora}
Taiqiang Wu, Jiahao Wang, Zhe Zhao, and Ngai Wong. 2024.
\newblock \href {https://doi.org/10.18653/v1/2024.emnlp-main.450} {Mixture-of-subspaces in low-rank adaptation}.
\newblock In \emph{Proceedings of the 2024 Conference on Empirical Methods in Natural Language Processing}, pages 7880--7899, Miami, Florida, USA. Association for Computational Linguistics.

\bibitem[{Xu et~al.(2019)Xu, Liu, Shu, and Yu}]{xu-etal-2019-bertpost}
Hu~Xu, Bing Liu, Lei Shu, and Philip Yu. 2019.
\newblock \href {https://doi.org/10.18653/v1/N19-1242} {{BERT} post-training for review reading comprehension and aspect-based sentiment analysis}.
\newblock In \emph{Proceedings of the 2019 Conference of the North {A}merican Chapter of the Association for Computational Linguistics: Human Language Technologies, Volume 1 (Long and Short Papers)}, pages 2324--2335, Minneapolis, Minnesota. Association for Computational Linguistics.

\end{thebibliography}

\appendix

\section{Datasets}
\label{sec:appendix}
\subsection{Continual DAP datasets}
\label{section:datasets:dap datasets}
\noindent\textbf{Statistics.}
We utilized the same datasets from \citet{ke2023continual}. For domain-adaptive pre-training stage, there are 6 unlabeled domain datasets, which is related to reviews and academic papers. The detailed statistic is in Table~\ref{tab:DAP dataset_summary}. 

For end-task classification datasets, please refer to Table ~\ref{tab:End task dataset_summary}. We have 6 end task datasets corresponding to the 6 domain data corpus. 
\begin{table}[ht]
    \centering
    \caption{Domain-specific corpora used for unlabeled domain-adaptive pre-training.}
    \label{tab:DAP dataset_summary}
    \resizebox{\columnwidth}{!}{
    \begin{tabular}{llc}
        \toprule
        \multicolumn{3}{c}{\textbf{Unlabeled Domain Datasets}} \\
        \midrule
        \textbf{Source} & \textbf{Dataset} & \textbf{Size (MB)} \\
        \midrule
        \multirow{3}{*}{Reviews} & Yelp Restaurant & 758 \\
        & Amazon Phone & 724 \\
        & Amazon Camera & 319 \\
        \midrule
        \multirow{3}{*}{Academic Papers} & ACL Papers & 867 \\
        & AI Papers & 507 \\
        & PubMed Papers & 989 \\
        \bottomrule
    \end{tabular}
    }
\end{table}
\begin{table}[ht]
    \centering
    \caption{Labeled datasets for end-task classification corresponding to each domain.}
    \label{tab:End task dataset_summary}
    \resizebox{\columnwidth}{!}{
    \begin{tabular}{llccc}
        \toprule
        \multicolumn{5}{c}{\textbf{End-Task Classification Datasets}} \\
        \midrule
        \textbf{Dataset} & \textbf{Task} & \textbf{\# Training} & \textbf{\# Testing} & \textbf{\# Classes} \\
        \midrule
        Restaurant & Aspect Sentiment Classification (ASC) & 3,452 & 1,120 & 3 \\
        Phone & Aspect Sentiment Classification (ASC) & 239 & 553 & 2 \\
        Camera & Aspect Sentiment Classification (ASC) & 230 & 626 & 2 \\
        ACL & Citation Intent Classification & 1,520 & 421 & 6 \\
        AI & Relation Classification & 2,260 & 2,388 & 7 \\
        PubMed & Chemical-Protein Interaction Prediction & 2,667 & 7,398 & 13 \\
        \bottomrule
    \end{tabular}
    }
\end{table}

\noindent\textbf{Examples.} We provide examples of each dataset and their corresponding classification tasks in Tables \ref{tab:yelp_reviews}, \ref{tab:rest_asc_examples}, \ref{tab:acl_papers}, \ref{tab:acl_citation}, \ref{tab:amazon_phone_reviews}, \ref{tab:phone_asc_examples}, \ref{tab:amazon_camera_reviews}, \ref{tab:camera_asc_examples}, \ref{tab:ai_papers_abstracts}, \ref{tab:ai_relation_classification}, \ref{tab:pubmed_papers_abstracts}, and \ref{tab:pubmed_cpip_examples}. Below, we provide detailed explanations of the end-task classifications:

1.	\textbf{Aspect Sentiment Classification (ASC)}:
This task is relevant for the Phone, Camera, and Restaurant datasets. The objective is to determine the sentiment (positive, negative, or neutral) associated with a specific aspect or feature mentioned in a review sentence.
	
2.	\textbf{Citation Intent Classification}:
This task is associated with the ACL dataset and involves analyzing sentences containing citations. The goal is to classify the function of the citation in the sentence into one of several predefined categories, such as “background,” “motivation,” “uses,” “extension,” “comparison or contrast,” or “future work.”
	
3.	\textbf{Relation Classification}:
Applied to the AI dataset, this task focuses on identifying the type of relationship expressed within a specified span of words that contains two entities in a sentence. The relationships to be classified include “feature of,” “conjunction,” “evaluate for,” “hyponym of,” “used for,” “part of,” and “compare.”
	
4.	\textbf{Chemical-Protein Interaction Classification}:
This task is designed for the PubMed dataset and involves determining the type of interaction between a chemical and a protein mentioned within a text span. The possible interaction types include “downregulator,” “substrate,” “indirect-upregulator,” “indirect-downregulator,” “agonist,” “activator,” “product of,” “agonist-activator,” “inhibitor,” “upregulator,” “substrate product of,” “agonist-inhibitor,” and “antagonist.”

\subsection{Reasoning datasets}
\noindent\textbf{Common-sense Reasoning.}
We follow the dataset of \citet{hu2023llmadapter}, so the dataset used in common-sense reasnoning task is represented in Table \ref{tab:cs_datasets}. From these 8 datasets, \citet{hu2023llmadapter} constructed \textit{Common-sense170K} and we utilized this dataset. We provide some examples of these dataset, in Table \ref{tab:boolq_examples}, \ref{tab:piqa_examples}, \ref{tab:arc_challenge_examples}.

\noindent\textbf{Arithemetic Reasoning.} In arithemetic task, we also follow the dataset of \citet{hu2023llmadapter}. The datset statistics is represented in Table \ref{tab:math_datasets}. \citet{hu2023llmadapter} constructed \textit{Math10k} and we utilized this datset. We provide some examples of these datasets in Table \ref{tab:math10k_examples}.

\begin{table}[ht]
    \centering
    \caption{Details of Common-Sense Reasoning Datasets}
    \label{tab:cs_datasets}
    \resizebox{\columnwidth}{!}{
    \begin{tabular}{lccc}
        \toprule
        \textbf{Dataset} & \textbf{\# Train} & \textbf{\# Test} & \textbf{Answer Format} \\
        \midrule
        BoolQ & 9.4K & 3,270 & Yes/No \\
        PIQA & 16.1K & 1,830 & Option \\
        SIQA & 33.4K & 1,954 & Option \\
        HellaSwag & 39.9K & 10,042 & Option \\
        WinoGrande & 63.2K & 1,267 & Option \\
        ARC-e & 1.1K & 2,376 & Option \\
        ARC-c & 2.3K & 1,172 & Option \\
        OBQA & 5.0K & 500 & Option \\
        \bottomrule
    \end{tabular}
    }
\end{table}
\begin{table}[ht]
    \centering
    \caption{Details of Arithemetic Reasoning Datasets}
    \label{tab:math_datasets}
    \resizebox{\columnwidth}{!}{
    \begin{tabular}{lccc}
        \toprule
        \textbf{Dataset} & \textbf{\# Train} & \textbf{\# Test} & \textbf{Answer Format} \\
        \midrule
        GSM8K & 8.8K & 1,319 & Number \\
        AQuA & 100K & 254 & Option \\
        \bottomrule
    \end{tabular}
    }
\end{table}

\section{Experimental Details}

This section provides a detailed explanation of the experimental setup. All experiments were conducted using four NVIDIA A100 GPUs with 80GB of memory each.

\subsection{Continual DAP Setting}
\label{section:appendix:DAPDetail}
\noindent\textbf{Architecture.}
For all continual DAP experiments, we adopt the \textit{RoBERTa-Base} model \cite{liu2019roberta}, following the settings described in \citet{ke2023continual}. During the DAP stage, a masked language modeling head is attached to the backbone model. In the end-task fine-tuning stage, a classification head, with outputs corresponding to the number of classes, is appended to the domain-adaptive pre-trained backbone.

\noindent\textbf{Hyperparameters.}
We primarily follow the hyperparameter settings of \citet{ke2023continual}, except for modifications explicitly mentioned here. During the DAP stage, we did not set the \textit{max samples} parameter, allowing the model to train on all samples from the domain dataset. In contrast, \citet{ke2023continual} set \textit{max samples} to 640,000, resulting in training for 2.5K steps per domain. This approach leads to underutilization of data in some domains and overutilization in others. Our setting is more realistic as it ensures equal utilization of domain data without overfitting, even for domains with fewer samples.

Additionally, for end-task fine-tuning, we conducted experiments using 10 seeds, compared to 5 seeds in \citet{ke2023continual}, for more robust results. We set the number of fine-tuning epochs to 15 and the learning rate to 3e-5. For the DAP stage with LoRA, the learning rate was set to 5e-4 and the batch size to 64. For the end-task fine-tuning stage with LoRA, the learning rate was set to 3e-4. The complete set of hyperparameters is detailed in Table \ref{tab:DAPhyperparameters}.
\begin{table*}[ht]
    \centering
    \caption{DAP Hyperparameter Settings}
    \label{tab:DAPhyperparameters}
    \resizebox{\textwidth}{!}{
    \begin{tabular}{ll}
        \toprule
        \textbf{Category} & \textbf{Hyperparameter} \\
        \midrule
        \textbf{General} & \\
        \quad Optimizer & AdamW \\
        \quad Max Sequence Length & 164 \\
        \quad Training Steps per Domain & All samples (full pass through domain data) \\
        \quad End-Task Fine-Tuning Epochs & 15 (all datasets) \\
        \quad \# of End-Task Fine-Tuning Seeds & 10 \\

        \midrule
        \textbf{DAP Stage} & \\
        \quad Learning Rate & 1e-4 \\
        \quad Batch Size & 256 \\
        \midrule
        \textbf{DAP Stage (With LoRA)} & \\
        \quad Learning Rate & 5e-4 \\
        \quad Batch Size & 64 \\
        \quad Target & All Linear Layer \\
        \quad Rank $r$ & 8 \\
        \quad $\alpha$ & 16 \\
        \midrule
        \textbf{End-Task Fine-Tuning Stage (Full-Finetuning)} & \\
        \quad Learning Rate & 3e-5 \\
        \quad Batch Size & 16 \\
        \midrule
        \textbf{End-Task Fine-Tuning Stage (With LoRA)} & \\
        \quad Learning Rate & 3e-4,     5e-5 \\
        \quad Batch Size & 16 \\
        \quad Target & All Linear Layer \\
        \quad Rank $r$ & 8, 48 \\
        \quad $\alpha$ & 16, 96 \\

        \bottomrule
    \end{tabular}
    }
\end{table*}

\noindent\textbf{Domain order.}  
We experimented with the following six domain orders, each ending with a different domain, to evaluate the robustness of methods under varying data orders:

\begin{enumerate}
    \item Yelp Restaurant $\rightarrow$ ACL Papers $\rightarrow$ AI Papers $\rightarrow$ Amazon Phone $\rightarrow$ PubMed Papers $\rightarrow$ Amazon Camera
    \item Amazon Phone $\rightarrow$ AI Papers $\rightarrow$ PubMed Papers $\rightarrow$ Amazon Camera $\rightarrow$ Yelp Restaurant $\rightarrow$ ACL Papers
    \item PubMed Papers $\rightarrow$ Amazon Camera $\rightarrow$ ACL Papers $\rightarrow$ Yelp Restaurant $\rightarrow$ AI Papers $\rightarrow$ Amazon Phone
    \item Amazon Camera $\rightarrow$ ACL Papers $\rightarrow$ Amazon Phone $\rightarrow$ Yelp Restaurant $\rightarrow$ PubMed Papers $\rightarrow$ AI Papers
    \item AI Papers $\rightarrow$ PubMed Papers $\rightarrow$ Amazon Camera $\rightarrow$ Amazon Phone $\rightarrow$ ACL Papers $\rightarrow$ Yelp Restaurant
    \item AI Papers $\rightarrow$ ACL Papers $\rightarrow$ Yelp Restaurant $\rightarrow$ Amazon Camera $\rightarrow$ Amazon Phone $\rightarrow$ PubMed Papers
\end{enumerate}

\subsection{Reasoning}
\label{section:appendix:hyperparams reasoning}
We utilize both LLaMA3-8B \citep{llama3modelcard} and Gemma2-9B \citep{team2024gemma} for the reasoning tasks. For LLaMA3-8B, we follow the hyperparameter settings reported in the respective paper. For Gemma2-9B, due to resource constraints, we reduce the batch size by half and accordingly lower the learning rate to ensure stable optimization. The full hyperparameter configurations are provided in Table~\ref{tab:Reasoning_hyperparameters}.

\begin{table*}[ht]
    \centering
    \caption{Hyperparameters used for reasoning tasks with LLaMA3-8B and Gemma2-9B.}
    \label{tab:Reasoning_hyperparameters}
    \resizebox{0.95\textwidth}{!}{
    \begin{tabular}{lcc}
        \toprule
        \textbf{Hyperparameter} & \textbf{LLaMA3-8B} & \textbf{Gemma2-9B} \\
        \midrule
        \textbf{Rank $r$} & 16, 32 & 16 \\
        \textbf{$\alpha$} & 32, 64 & 32 \\
        \textbf{Dropout} & 0.05 & 0.05 \\
        \textbf{Optimizer} & AdamW & AdamW \\
        \textbf{Learning Rate (Common-sense)} & 1e-4 & 5e-5 \\
        \textbf{Learning Rate (Arithmetic)} & 3e-4 & 1.5e-4 \\
        \textbf{Learning Rate Scheduler} & Linear & Linear \\
        \textbf{Batch Size} & 16 & 8 \\
        \textbf{Sequence Length} & 256 & 256 \\
        \textbf{Warmup Steps} & 100 & 100 \\
        \textbf{Epochs} & 3 & 3 \\
        \textbf{Target Modules} & Q, K, V, Up, Down & Q, K, V, Up, Down \\
        \bottomrule
    \end{tabular}
    }
\end{table*}

\section{Details of Experiments}
\subsection{Details of the Motivation Experiment}
\label{section: Appendix: DetailofMotivation}
The exact domain names presented in Figure~\ref{fig:MotivationHeatmap}, in the same order, are listed in Table~\ref{tab:DAPResults}. The DAP stage follows the same procedure as \textbf{Separate LoRA} in Table~\ref{tab:DAPResults}. Fine-tuning is conducted using full fine-tuning, with hyperparameters specified in Table~\ref{tab:DAPhyperparameters}.


\subsection{Details of GPU Memory Overhead According to the Number of Domains}
\label{section: Appendix: numDomainGPU}

In Table~\ref{tab:numparam}, we report the number of trainable parameters and the GPU memory usage during end-task fine-tuning for various settings. As shown, although GPU memory usage increases linearly with the number of domains, the growth is minimal. This implies that DoMIX can scale to more domains without significant overhead.

\begin{figure}[t]
  \includegraphics[width=\columnwidth]{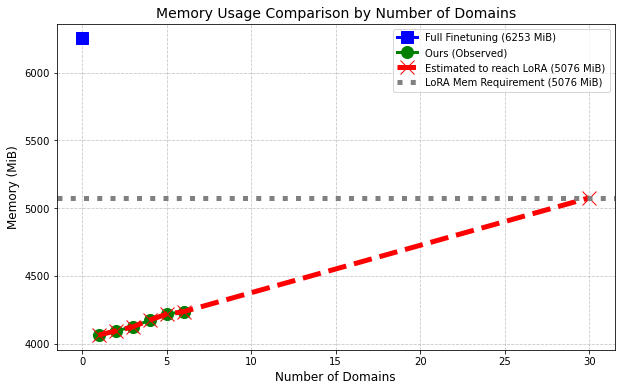}
  \caption{Comparison of memory usage by increasing number of domains.}
  \label{fig:memcostgraph}
\end{figure}

Figure~\ref{fig:memcostgraph} visualizes this trend and includes an extrapolated estimate of memory usage as the number of domains increases. The dotted gray line indicates the GPU memory usage of standard LoRA with rank 48. Note that this rank corresponds to the total parameter size of DoMIX when using 6 domains with rank 8 each—excluding the bridge matrix. Despite having a similar number of trainable parameters, DoMIX consumes significantly less GPU memory due to its partially frozen structure and selective training. The extrapolated red line shows that only when scaling to around 30 domains does DoMIX reach the same memory usage as standard LoRA ($r=48$), highlighting its strong scalability and memory efficiency.

This scalability is possible because DoMIX freezes the input-side LoRA matrices (A) and selectively trains only the necessary components, thereby minimizing memory overhead per added domain.

\subsection{Detailed Results of Ablation Studies}
In this section, we provide task-wise results for the ablation studies, complementing the average accuracy and F1 scores reported in the main text. 
Tables~\ref{tab:ablation_P_detail}, \ref{tab:ablation_freezing_detail}, and \ref{tab:ablation_initialization_detail} respectively provide the detailed results corresponding to Tables~\ref{tab:ablation_P}, \ref{tab:ablation_freezing}, and \ref{tab:ablation_initialization}.

\begin{table*}[t]
\centering
\caption{Task-wise results, comparing the presence and trainability of the bridge module $P$.
}
\resizebox{\textwidth}{!}{
\begin{tabular}{lc|ccccccccccc|cc}
\toprule
\textbf{\multirow{2}{*}{$P$ init.}} & \textbf{\multirow{2}{*}{Trainable}} 
& \multicolumn{2}{c}{\textbf{Restaurant}} 
& \multicolumn{2}{c}{\textbf{ACL}} 
& \multicolumn{2}{c}{\textbf{AI}} 
& \multicolumn{2}{c}{\textbf{Phone}} 
& \textbf{PubMed}
& \multicolumn{2}{c|}{\textbf{Camera}} 
& \multicolumn{2}{c}{\textbf{Avg.}} \\
& 
& Acc & F1 & Acc & F1 & Acc & F1 & Acc & F1 & F1 & Acc & F1 & Acc & F1 \\
\midrule
\multirow{2}{*}{Id.} & \xmark 
& 82.89 & 71.53 & 71.16 & 66.71 & 51.68 & 30.54 & 75.48 & 66.84 & 62.17 & 85.45 & 74.16 & 71.47 & 61.99 \\
 & \cmark 
& 83.67 & 74.20 & 67.91 & 61.17 & 67.12 & 55.33 & 76.06 & 66.88 & 61.94 & 84.95 & 73.13 & 73.61 & 65.44 \\
\midrule
\multirow{2}{*}{Ours} & \xmark 
& 85.09 & 75.63 & 57.89 & 40.20 & 64.10 & 48.10 & 64.38 & 39.16 & 68.64 & 77.48 & 43.65 & 69.59 & 52.57 \\
& \cmark 
& \textbf{86.67} & \textbf{79.30} & \textbf{72.97} & \textbf{69.10} & \textbf{79.11} & \textbf{74.01} & \textbf{87.12} & \textbf{85.23} & \textbf{73.61} & \textbf{90.54} & \textbf{85.79} & \textbf{81.67} & \textbf{77.84} \\
\bottomrule
\end{tabular}
}
\label{tab:ablation_P_detail}
\end{table*}

\begin{table*}[t]
\centering
\caption{Task-wise results for different trainable configurations of $A$, $P$, and $B$.}
\resizebox{\textwidth}{!}{
\begin{tabular}{ccc|ccccccccccc|cc}
\toprule
\multicolumn{3}{c|}{\textbf{Trainable}} 
& \multicolumn{2}{c}{\textbf{Restaurant}} 
& \multicolumn{2}{c}{\textbf{ACL}} 
& \multicolumn{2}{c}{\textbf{AI}} 
& \multicolumn{2}{c}{\textbf{Phone}} 
& \textbf{PubMed}
& \multicolumn{2}{c|}{\textbf{Camera}} 
& \multicolumn{2}{c}{\textbf{Avg.}} \\
\textbf{A} & \textbf{P} & \textbf{B} 
& Acc & F1 & Acc & F1 & Acc & F1 & Acc & F1 & F1 & Acc & F1 & Acc & F1 \\
\midrule
\cmark & \xmark & \cmark 
& 86.67 & \textbf{79.47} & 71.47 & 67.06 & 77.42 & 71.76 & \textbf{87.96} & \textbf{86.32} & 72.47 & \textbf{92.38} & \textbf{88.73} & 81.39 & 77.63 \\
\cmark & \cmark & \xmark 
& \textbf{86.75} & 79.46 & 71.43 & 66.75 & 77.04 & 71.04 & 86.93 & 84.94 & 72.19 & 90.13 & 85.06 & 80.74 & 76.57 \\
\cmark & \cmark & \cmark 
& 84.96 & 74.89 & 72.30 & 68.65 & \textbf{79.60} & \textbf{74.57} & 86.87 & 84.87 & \textbf{73.86} & 90.24 & 85.17 & 81.31 & 77.00 \\
\xmark & \cmark & \cmark 
& 86.67 & 79.30 & \textbf{72.97} & \textbf{69.10} & 79.11 & 74.01 & 87.12 & 85.23 & 73.61 & 90.54 & 85.79 & \textbf{81.67} & \textbf{77.84} \\
\bottomrule
\end{tabular}
}
\label{tab:ablation_freezing_detail}
\end{table*}

\begin{table*}[t]
\centering
\caption{Task-wise results for different initialization strategies of $A$, $P$, and $B$.}
\resizebox{\textwidth}{!}{
\begin{tabular}{ccc|ccccccccccc|cc}
\toprule
\multicolumn{3}{c|}{\textbf{Initialization}} 
& \multicolumn{2}{c}{\textbf{Restaurant}} 
& \multicolumn{2}{c}{\textbf{ACL}} 
& \multicolumn{2}{c}{\textbf{AI}} 
& \multicolumn{2}{c}{\textbf{Phone}} 
& \textbf{PubMed}
& \multicolumn{2}{c|}{\textbf{Camera}} 
& \multicolumn{2}{c}{\textbf{Avg.}} \\
\textbf{A} & \textbf{P} & \textbf{B} 
& Acc & F1 & Acc & F1 & Acc & F1 & Acc & F1 & F1 & Acc & F1 & Acc & F1 \\
\midrule
Ours & Zero    & Ours 
& \textbf{87.30} & \textbf{80.11} & \textbf{73.47} & \textbf{69.94} & 78.04 & 71.76 & 85.01 & 82.67 & 73.15 & 90.67 & 86.16 & 81.27 & 77.30 \\
Ours & Kaiming & Ours 
& 86.21 & 78.75 & 71.69 & 67.34 & 77.26 & 70.84 & 83.73 & 80.40 & 71.93 & 89.81 & 84.22 & 80.10 & 75.58 \\
Ours & Ours    & Zero 
& 86.41 & 78.80 & 71.76 & 67.59 & 78.14 & 72.21 & 86.67 & 84.55 & 73.02 & 90.78 & 86.03 & 81.13 & 77.04 \\
Ours & Ours    & Kaiming 
& 86.63 & 79.43 & 71.78 & 67.53 & 77.70 & 71.40 & 86.67 & 84.62 & 72.48 & \textbf{91.41} & \textbf{87.30} & 81.11 & 77.12 \\
Kaiming & Ours & Ours 
& 86.13 & 78.68 & 71.24 & 66.64 & 77.89 & 72.45 & \textbf{87.92} & \textbf{86.40} & 72.52 & 91.18 & 87.06 & 81.15 & 77.29 \\
Ours & Ours    & Ours 
& 86.67 & 79.30 & 72.97 & 69.10 & \textbf{79.11} & \textbf{74.01} & 87.12 & 85.23 & \textbf{73.61} & 90.54 & 85.79 & \textbf{81.67} & \textbf{77.84} \\
\bottomrule
\end{tabular}
}
\label{tab:ablation_initialization_detail}
\end{table*}

\section{Ablation Study on Bridge Module $P$ in LLaMA3-8B}
In this section, we present ablation experiments on the LLaMA3-8B model to assess the effects of the presence and trainability of the bridge module $P$ (corresponding to Table~\ref{tab:ablation_P} in Section~\ref{section:ablation:effectofP}).
The results are shown in Table~\ref{tab:ablation_p_llama}, and they are consistent with those reported in Table~\ref{tab:ablation_P}, highlighting the necessity of the bridge module and the importance of its trainability—even in large-scale LLMs.

\begin{table*}[t]
\centering
\caption{Task-wise commonsense reasoning results on LLaMA3-8B, evaluating the impact of the presence and trainability of the bridge module $P$.}
\resizebox{\textwidth}{!}{
\begin{tabular}{lc|cccccccc|c}
\toprule
\textbf{{$P$ init.}} & \textbf{{Trainable}} 
& \textbf{WinoG.} & \textbf{OBQA} & \textbf{SIQA} & \textbf{ARC-e} & \textbf{BoolQ} 
& \textbf{ARC-c} & \textbf{PIQA} & \textbf{HellaS.} 
& \textbf{Avg.} \\
\midrule
\multirow{2}{*}{Id.} & \xmark 
& 84.77 & 83.60 & 80.25 & 86.45 & 72.11 & 75.68 & 87.54 & 94.11 & 83.06 \\
& \cmark 
& 84.53 & 84.20 & 80.19 & 86.62 & 72.14 & 76.02 & 87.49 & 94.14 & 83.17 \\
\midrule
\multirow{2}{*}{Ours} & \xmark 
& 75.06 & 77.00 & 72.11 & 88.80 & 22.91 & 73.72 & 86.02 & 89.02 & 73.08 \\
& \cmark 
& \textbf{85.95} & \textbf{85.80} & \textbf{80.25} & \textbf{91.04} & \textbf{73.94} 
& \textbf{80.55} & \textbf{88.36} & \textbf{95.85} & \textbf{85.22} \\
\bottomrule
\end{tabular}
}
\label{tab:ablation_p_llama}
\end{table*}

\begin{table*}[ht]
    \centering
    \caption{Examples of Yelp Restaurant Reviews}
    \label{tab:yelp_reviews}
    \resizebox{\textwidth}{!}{
    \begin{tabular}{lp{0.9\textwidth}}
        \toprule
        \textbf{ID} & \textbf{Review Text} \\
        \midrule
        1 & The food is always great here. The service from both the manager as well as the staff is super. Only drawback of this restaurant is it's super loud. If you can, snag a patio table! \\
        2 & This place used to be a cool, chill place. Now it's a bunch of neanderthal bouncers hopped up on steroids acting like they can do whatever they want. There are so many better places in Davis Square where they are glad you are visiting their business. Sad that the Burren is now the worst place in Davis. \\
        3 & The setting is perfectly adequate, and the food comes close. The dining chains like Chili's and Victoria Station do barbecue better. It's no surprise you can always pick up coupons for Linwood at restaurant.com. \\
        \bottomrule
    \end{tabular}
    }
\end{table*}

\begin{table*}[ht]
    \centering
    \caption{Examples of Restaurant Aspect Sentiment Classification (ASC)}
    \label{tab:rest_asc_examples}
    \resizebox{\textwidth}{!}{
    \begin{tabular}{lllp{0.7\textwidth}}
        \toprule
\textbf{ID} & \textbf{Polarity} & \textbf{Aspect Term} & \textbf{Sentence} \\
        \midrule
        0 & Positive & Staff & The staff is very kind and well-trained. They're fast, they are always prompt to jump behind the bar and fix drinks, they know details of every item on the menu, and make excellent recommendations. \\
        1 & Negative & Server & The service is always bad though, don't expect much of anything from your server, and I would not recommend bringing a date here either. \\
        2 & Neutral & Brunch & Where tanks in other Chinatown restaurants display a lurking myriad of sad-looking marine life in their murky waters, the tanks at Ping's are clear as glass with healthy-looking creatures who do not yet know that they will be part of some dim sum lover's brunch. \\
        3 & Positive & Food & This was my first time at Cafe St. Bart's, and I must say how delicious the food and the service was. \\
        \bottomrule
    \end{tabular}
    }
\end{table*}

\begin{table*}[ht]
    \centering
    \caption{Examples of ACL Papers}
    \label{tab:acl_papers}
    \resizebox{\textwidth}{!}{
    \begin{tabular}{lp{0.9\textwidth}}
        \toprule
        \textbf{ID} & \textbf{Text} \\
        \midrule
        1 & This paper describes the three phases of the Durkheim Project. For this project we developed a clinician's dashboard that displays output of models predicting suicide risk of veterans and active duty military personnel. During phase one, we built the clinician's dashboard and completed a Veterans Affairs (VA) predictive risk medical records study, based on an analysis of the narrative, or free text, portions of VA medical records. In phase two, we will predict suicide risk based on opt-in social media postings by patients using social media websites, e.g., Facebook. We describe the software infrastructure that we have completed for this phase two system. During phase three we will provide a three-layer intervention strategy. We discuss our methodology for the three phases, including IRB-approved protocols for the first two phases and a soon-to-be approved IRB protocol for phase three. \\
        2 & Diagnosis of psychological health and the prediction of negative events, such as suicide, or suicide ideation, is limited by: a) a lack of understanding of the true differentiating risks of suicidality (Health Promotion, 2010; Treating Soldiers, 2010) and b) a lack of near real-time reaction capability to large volumes of data. There is a need for broader coverage suicide risk detection and a better understanding of the expression of suicide ideation through data mining of text and images. The Durkheim Project's proposed solution is to provide continuous monitoring of text-based information, such as found in social network user behavioral intent enabling intervention; facilitated by social/online data sources, powered by a medically-validated suicide risk classifier. \\
        \bottomrule
    \end{tabular}
    }
\end{table*}

\begin{table*}[ht]
    \centering
    \caption{Examples of ACL Citation Intent Classification}
    \label{tab:acl_citation}
    \resizebox{\textwidth}{!}{
    \begin{tabular}{lp{0.9\textwidth}l}
        \toprule
        \textbf{ID} & \textbf{Sentence} & \textbf{Label} \\
        \midrule
        1 & This was done by MERT optimization (Och, 2003) towards post-edits under the TER target metric. & Uses \\
        2 & She evaluates 3,000 German verbs with a token frequency between 10 and 2,000 against the Duden (Dudenredaktion 2001). & Background \\
        3 & The following four components have been identified as the key elements of a question related to patient care (Richardson et al. 1995): & Background \\
        4 & Briscoe and Carroll (1997) report on manually analyzing an open-class vocabulary of 35,000 head words for predicate subcategorization information and comparing the results against the subcategorization details in COMLEX. & CompareOrContrast \\
        \bottomrule
    \end{tabular}
    }
\end{table*}

\begin{table*}[ht]
    \centering
    \caption{Examples of Amazon Phone Reviews}
    \label{tab:amazon_phone_reviews}
    \resizebox{\textwidth}{!}{
    \begin{tabular}{lp{0.9\textwidth}}
        \toprule
        \textbf{ID} & \textbf{Review Text} \\
        \midrule
        1 & Saw this same case at a theme park store for 25 dollars. This is very good quality for a great price. \\
        2 & Best phone case ever. Everywhere I go I get a ton of compliments on it. It was in perfect condition as well. \\
        3 & The case is good, but the two pieces do not fit all the way together. It is slightly off and no matter how much I tried (even shaving the side a little) it wouldn't slide into each other. The gap is very, very little and you can barely notice it. The bottom of the case has never slid off, so I will deal with it for the wonderful price! \\
        \bottomrule
    \end{tabular}
    }
\end{table*}

\begin{table*}[ht]
    \centering
    \caption{Examples of Phone Aspect Sentiment Classification (ASC)}
    \label{tab:phone_asc_examples}
    \resizebox{\textwidth}{!}{
    \begin{tabular}{lllp{0.7\textwidth}}
        \toprule
        \textbf{ID} & \textbf{Polarity} & \textbf{Aspect Term} & \textbf{Sentence} \\
        \midrule
        0 & Positive & Work & There is much which has been said in other reviews about the features of this phone, it is a great phone, mine worked without any problems right out of the box. \\
        1 & Positive & Phone & There is much which has been said in other reviews about the features of this phone, it is a great phone, mine worked without any problems right out of the box. \\
        2 & Negative & AT\&T Customer Service & After several years of torture in the hands of AT\&T customer service, I am delighted to drop them, and look forward to August 2004 when I will convert our other 3 family-phones from AT\&T to T-Mobile! \\
        3 & Positive & Signal Quality & I have had the phone for 1 week, the signal quality has been great in the Detroit area (suburbs) and in my recent road trip between Detroit and northern Kentucky (Cincinnati) I experienced perfect signal and reception along I-75, far superior to AT\&T's which does not work along several long stretches on that same route. \\
        4 & Positive & Speaker Phone, Radio, Infrared & My favorite features, although there are many, are the speaker phone, the radio, and the infrared. \\
        \bottomrule
    \end{tabular}
    }
\end{table*}

\begin{table*}[ht]
    \centering
    \caption{Examples of Amazon Camera Reviews}
    \label{tab:amazon_camera_reviews}
    \resizebox{\textwidth}{!}{
    \begin{tabular}{lp{0.9\textwidth}}
        \toprule
        \textbf{ID} & \textbf{Review Text} \\
        \midrule
        1 & I bought this battery as a backup for my camera. It seems to work okay, but even "fully charged," the camera only reads the battery as 75\% full. I haven't used it enough to know if the battery life really is shorter than expected, though. \\
        2 & This battery worked great in my camera. My old battery was lasting less than a day. This battery took me through over half of my vacation with lots of use before I needed to recharge it. Great deal! \\
        3 & Why pay 3x the amount for an original equipment Sony battery when this one works just as well? Great product!! \\
        4 & I purchased two of these for my Sony digital point-and-shoot camera because the original ones died. They work very well and are listed at a good price, certainly much less than the replacement Sony batteries. \\
        \bottomrule
    \end{tabular}
    }
\end{table*}

\begin{table*}[ht]
    \centering
    \caption{Examples of Camera Aspect Sentiment Classification (ASC)}
    \label{tab:camera_asc_examples}
    \resizebox{\textwidth}{!}{
    \begin{tabular}{llll}
        \toprule
        \textbf{ID} & \textbf{Polarity} & \textbf{Aspect Term}& \textbf{Sentence} \\
        \midrule
        0 & Positive & Canon Powershot G3 & I recently purchased the Canon Powershot G3 and am extremely satisfied with the purchase. \\
        1 & Positive & Use & The camera is very easy to use. In fact, on a recent trip this past week, I was asked to take a picture of a vacationing elderly group. \\
        2 & Positive & Picture & They fired away, and the picture turned out quite nicely (as all of my pictures have thus far). \\
        3 & Positive & Picture Quality &A few of my work constituents owned the G2 and highly recommended the Canon for picture quality. \\
        4 & Positive & Picture Quality & I'm easily enlarging pictures to 8 1/2 x 11 with no visible loss in picture quality and not even using the best possible setting yet (super fine). \\
        \bottomrule
    \end{tabular}
    }
\end{table*}

\begin{table*}[ht]
    \centering
    \caption{Examples of AI Papers}
    \label{tab:ai_papers_abstracts}
    \resizebox{\textwidth}{!}{
    \begin{tabular}{lp{0.9\textwidth}}
        \toprule
        \textbf{ID} & \textbf{Text} \\
        \midrule
        1 & We consider two novel representations and feature extraction schemes for automatic recognition of emotion-related facial expressions. In one scheme, facial landmark points are tracked over successive video frames using an effective detector and tracker to extract landmark trajectories. Features are extracted from landmark trajectories using the Independent Component Analysis (ICA) method. In the alternative scheme, the evolution of the emotion expression on the face is captured by stacking normalized and aligned faces into a spatiotemporal face cube. Emotion descriptors are then 3D Discrete Cosine Transform (DCT) features from this prism or DCT \& ICA features. Several classifier configurations are used, and their performance is determined in detecting the 6 basic emotions. Decision fusion applied to classifiers improved the recognition performance of the best classifier by 9 percentage points. \\
        2 & The human face is a rich source of nonverbal information. Indeed, not only is it the source of identity information, but it also provides clues to understanding social feelings and can reveal mental states via social signals. Facial expressions form a significant part of human social interaction. Automatic understanding of emotions from face images is instrumental in the design of affective human-computer interfaces. Next-generation human-computer interfaces will be empowered with the capability to recognize and respond to nonverbal communication clues. \\
        3 & In this study, we consider two types of data representation for emotion analysis, the first one being facial landmark trajectories, and the second one being the evolution of face texture patches. Discriminative features are extracted from these two face representations for automatic facial expression recognition. Based on these features, we develop a novel algorithm for the automatic classification of emotional expressions in the face. \\
        \bottomrule
    \end{tabular}
    }
\end{table*}

\begin{table*}[ht]
    \centering
    \caption{Examples of AI Relation Classification}
    \label{tab:ai_relation_classification}
    \resizebox{\textwidth}{!}{
    \begin{tabular}{lp{0.7\textwidth}l}
        \toprule
        \textbf{ID} & \textbf{Sentence} & \textbf{Label} \\
        \midrule
        1 & The agreement in question involves number in [[nouns]] and <<reflexive pronouns>> and is syntactic rather than semantic in nature because grammatical number in English, like grammatical gender in languages such as French, is partly arbitrary. & CONJUNCTION \\
        2 & The agreement in question involves number in nouns and reflexive pronouns and is syntactic rather than semantic in nature because grammatical number in English, like [[grammatical gender]] in <<languages>> such as French, is partly arbitrary. & FEATURE-OF \\
        3 & The agreement in question involves number in nouns and reflexive pronouns and is syntactic rather than semantic in nature because grammatical number in English, like grammatical gender in <<languages>> such as [[French]], is partly arbitrary. & HYPONYM-OF \\
        4 & In this paper, a novel [[method]] to learn the <<intrinsic object structure>> for robust visual tracking is proposed. & USED-FOR \\
        \bottomrule
    \end{tabular}
    }
\end{table*}

\begin{table*}[ht]
    \centering
    \caption{Examples of PubMed Papers}
    \label{tab:pubmed_papers_abstracts}
    \resizebox{\textwidth}{!}{
    \begin{tabular}{lp{0.9\textwidth}}
        \toprule
        \textbf{ID} & \textbf{Text} \\
        \midrule
        1 & Several gold(I) complexes containing a thiolate ligand functionalized with several amino acid or peptide moieties of the type [Au(SPyCOR)(PPh2R')] (where R = OH, amino acid or dipeptide and R' = Ph or Py) were prepared. These thiolate gold complexes bearing biological molecules possess potential use as antitumor agents. Cytotoxicity assays in different tumour cell lines such as A549 (lung carcinoma), Jurkat (T-cell leukaemia) and MiaPaca2 (pancreatic carcinoma) revealed that the complexes exhibit good antiproliferative activity, with IC50 values in the low micromolar range. Several structural modifications were carried out to establish the structure-activity relationship in this family of complexes, which has led to the design of new and more potent cytotoxic complexes. Observations of different cellular events after the addition of the complexes indicated the possible mechanism of action or the biological targets of this type of new gold(I) drug. \\
        2 & Mass spectrometry provides a versatile detection method for high-throughput drug screening because it permits the use of native biological substrates and the direct quantification of unlabeled reaction products. This paper describes the design and application of a Swan-shaped probe for high-throughput and nanoliter-scale analysis of biological samples in both a microfluidic droplet array and a multiwell plate with electrospray ionization mass spectrometry (ESI-MS). The Swan probe is fabricated using a single capillary and consists of a U-shaped section with a micrometer-sized hole for sampling and a tapered tip for sample electrospray ionization. To validate its potential in drug discovery, the present system was applied in the screening of inhibitors of acetylcholinesterase (AchE) and the measurement of the IC50 values of identified inhibitors. \\
        \bottomrule
    \end{tabular}
    }
\end{table*}

\begin{table*}[ht]
    \centering
    \caption{Examples of PubMed Chemical-Protein Interaction Prediction}
    \label{tab:pubmed_cpip_examples}
    \resizebox{\textwidth}{!}{
    \begin{tabular}{lp{0.7\textwidth}l}
        \toprule
        \textbf{ID} & \textbf{Sentence} & \textbf{Label} \\
        \midrule
        1 & Taken together, our results demonstrate the ability of <<Fc11a-2>> to inhibit [[NLRP3]] inflammasome activation and its potential use in the treatment of inflammatory bowel diseases. & INHIBITOR \\
        2 & At PND35, the medial prefrontal cortex (mPFC) of rats given <<MPH>> showed 55\% greater immunoreactivity (-ir) for the catecholamine marker [[tyrosine hydroxylase]] (TH), 60\% more Nissl-stained cells, and 40\% less norepinephrine transporter (NET)-ir density. & INDIRECT-UPREGULATOR \\
        3 & <<Epidermal growth factor receptor>> inhibitors currently under investigation include the small molecules gefitinib (Iressa, ZD1839) and erlotinib (Tarceva, OSI-774), as well as monoclonal antibodies such as cetuximab (IMC-225, [[Erbitux]]). & INHIBITOR \\
        4 & Agents that have only begun to undergo clinical evaluation include <<CI-1033>>, an irreversible pan-[[erbB]] tyrosine kinase inhibitor, and PKI166 and GW572016, both examples of dual kinase inhibitors (inhibiting epidermal growth factor receptor and Her2). & INHIBITOR \\
        \bottomrule
    \end{tabular}
    }
\end{table*}

\begin{table*}[ht]
    \centering
    \caption{Examples from the BoolQ Dataset}
    \label{tab:boolq_examples}
    \resizebox{\textwidth}{!}{
    \begin{tabular}{lp{\textwidth}l}
        \toprule
        \textbf{ID} & \textbf{Instruction and Question} & \textbf{Answer} \\
        \midrule
        1 & Please answer the following question with true or false: is the Golden State Warriors in the playoffs? \newline \textbf{Answer format: true/false} \newline \textbf{Output:} The correct answer is true. & True \\
        2 & Please answer the following question with true or false: Downton Abbey, will there be a season 7? \newline \textbf{Answer format: true/false} \newline \textbf{Output:} The correct answer is false. & False \\
        \bottomrule
    \end{tabular}
    }
\end{table*}

\begin{table*}[ht]
    \centering
    \caption{Examples from the PIQA Dataset}
    \label{tab:piqa_examples}
    \resizebox{\textwidth}{!}{
    \begin{tabular}{lp{0.9\textwidth}l}
        \toprule
        \textbf{ID} & \textbf{Instruction, Question, and Solutions} & \textbf{Answer} \\
        \midrule
        1 & \textbf{Instruction:} Please choose the correct solution to the question: To seal leather for furniture. \newline
        \textbf{Solution1:} Place the leather on a work surface, pour a small amount of Neatsfoot oil onto the leather, rub the oil into all parts of the leather with a sponge, and allow the leather to dry overnight. \newline
        \textbf{Solution2:} Place the leather on a work surface, pour a small amount of Neatsfoot oil onto the leather, rub the oil into all parts of the leather with your hands, and allow the leather to dry overnight. \newline
        \textbf{Answer format:} solution1/solution2 \newline
        \textbf{Output:} The correct answer is solution2. & Solution2 \\
        \midrule
        2 & \textbf{Instruction:} Please choose the correct solution to the question: Make lemonade. \newline
        \textbf{Solution1:} Fill a glass about 3/4 full with water. Using freshly squeezed lemon juice or bottled juice, add 1-2 tablespoons to the water. Sweeten with sugar or non-caloric sweetener to taste. Add ice and enjoy! \newline
        \textbf{Solution2:} Fill a glass about 3/4 full with water. Using freshly squeezed lemon juice or bottled juice, add 1-2 tablespoons to the water. Sweeten with sugar or non-caloric sweetener to taste. Add gin and enjoy! \newline
        \textbf{Answer format:} solution1/solution2 \newline
        \textbf{Output:} The correct answer is solution1. & Solution1 \\
        \bottomrule
    \end{tabular}
    }
\end{table*}

\begin{table*}[ht]
    \centering
    \caption{Examples from the ARC-Challenge Dataset}
    \label{tab:arc_challenge_examples}
    \resizebox{\textwidth}{!}{
    \begin{tabular}{lp{0.7\textwidth}l}
        \toprule
        \textbf{ID} & \textbf{Instruction, Question, and Answers} & \textbf{Answer} \\
        \midrule
        1 & \textbf{Instruction:} Please choose the correct answer to the question: Screech owls have two color variations—red and grey. What advantage does the grey screech owl have over the red screech owl in a habitat that is made up of trees with dark-colored bark? \newline
        \textbf{Answer1:} Nesting \newline
        \textbf{Answer2:} Feeding \newline
        \textbf{Answer3:} Reproduction \newline
        \textbf{Answer4:} Camouflage \newline
        \textbf{Answer format:} answer1/answer2/answer3/answer4 \newline
        \textbf{Output:} The correct answer is answer4. & Answer4 \\
        \midrule
        2 & \textbf{Instruction:} Please choose the correct answer to the question: A scientist is measuring the amount of movement along a fault. Which tool is best used for making this measurement? \newline
        \textbf{Answer1:} Barometer \newline
        \textbf{Answer2:} Stopwatch \newline
        \textbf{Answer3:} Meter stick \newline
        \textbf{Answer4:} Magnifying lens \newline
        \textbf{Answer format:} answer1/answer2/answer3/answer4 \newline
        \textbf{Output:} The correct answer is answer3. & Answer3 \\
        \bottomrule
    \end{tabular}
    }
\end{table*}

\begin{table*}[ht]
    \centering
    \caption{Examples from the Math10K Dataset}
    \label{tab:math10k_examples}
    \resizebox{\textwidth}{!}{
    \begin{tabular}{lp{0.9\textwidth}}
        \toprule
        \textbf{ID} & \textbf{Instruction, Reasoning, and Answer} \\
        \midrule
        1 & \textbf{Instruction:} Find the area of a circle whose radius is 70m? The options are: A) 28866, B) 23456, C) 76766, D) 16866, E) 15400. \newline
        \textbf{Reasoning:} The formula for the area of a circle is \(A = \pi r^2\), where \(r\) is the radius of the circle. Substituting the given radius, we get: \(A = \pi(70)^2 = 4900\pi\). Approximating \(\pi\) to 3.14 gives \(A \approx 15400\). Therefore, the answer is E) 15400. \newline
        \textbf{Answer:} E \\
        \midrule
        2 & \textbf{Instruction:} Jimmy has a collection of 5 action figures. Each figure is worth \$15, except for one which is worth \$20. He decided to sell his collection. To do it fast, he decided to sell each of them for \$5 less than their value. How much will Jimmy earn if he sells all the figures? \newline
        \textbf{Reasoning:} Step 1: Calculate the total value of the collection: 4 action figures are worth \$15 each (\$60), and 1 is worth \$20. Total value = \$60 + \$20 = \$80. \newline
        Step 2: Calculate the selling price: Each figure will be sold for \$5 less than its value. For 4 figures worth \$15 each: \$15 - \$5 = \$10. For the figure worth \$20: \$20 - \$5 = \$15. \newline
        Step 3: Calculate total earnings: Jimmy earns \$10 for 4 figures (\$40) and \$15 for 1 figure. Total earnings = \$40 + \$15 = \$55. \newline
        \textbf{Answer:} 55.0 \\
        \bottomrule
    \end{tabular}
    }
\end{table*}
\end{document}